\documentclass[12pt, a4paper]{article}

\usepackage[T1]{fontenc}
\usepackage[utf8]{inputenc}
\usepackage{newtxtext,newtxmath}
\usepackage[margin=2.5cm]{geometry}

\usepackage{amsmath, amssymb}
\usepackage{booktabs}
\usepackage{array}
\usepackage{longtable}
\usepackage{multirow}
\usepackage{xcolor}
\usepackage{hyperref}
\usepackage{natbib}
\usepackage{setspace}
\usepackage{parskip}
\usepackage{authblk}
\usepackage{titlesec}
\usepackage{enumitem}
\usepackage{graphicx}
\usepackage{subcaption}
\usepackage{caption}
\usepackage{placeins}

\hypersetup{
  colorlinks = true,
  linkcolor = blue!70!black,
  citecolor = blue!70!black,
  urlcolor = blue!70!black
}

\titleformat{\section}{\large\bfseries}{\thesection.}{0.6em}{}
\titleformat{\subsection}{\normalsize\bfseries}{\thesubsection}{0.6em}{}
\titleformat{\subsubsection}{\normalsize\itshape}{\thesubsubsection}{0.6em}{}

\begin{document}

\begin{center}

{\Large\bfseries
Seeing SDG 6 from space: local-scale monitoring of piped water and sewage systems across Africa using satellite imagery and self-supervised learning\par}

\vspace{3em}

{\normalsize
Othmane Echchabi$^{1,2,5, \dagger}$, Aya Lahlou$^{3,4,5,\dagger}$, Nizar Talty$^{5}$, Josh Malcolm Manto$^{5}$, Tongshu Zheng$^{5,*}$, and Ka Leung Lam$^{5,*}$\par}
\end{center}

\vspace{0.8em}

{\small
$^{1}$ Mila -- Quebec AI Institute\par
$^{2}$ School of Computer Science, McGill University\par
$^{3}$ Department of Earth and Environmental Engineering, Columbia University\par
$^{4}$ Center for Learning the Earth with Artificial Intelligence and Physics (LEAP)\par
$^{5}$ Division of Natural and Applied Sciences, Duke Kunshan University\par}


\vspace{0.5em}

{\footnotesize
$^{*}$ Corresponding authors: \href{mailto:tongshu.zheng@dukekunshan.edu.cn}{tongshu.zheng@dukekunshan.edu.cn}; \href{mailto:kaleung.lam@dukekunshan.edu.cn}{kaleung.lam@dukekunshan.edu.cn}\par
$^{\dagger}$ These authors contributed equally to this work.\par}

\vspace{1em}

\begin{abstract}
\noindent

Access to drinking water and sanitation services is essential. Sustainable Development Goal 6 (SDG 6) aims to achieve universal access, but progress monitoring remains constrained by costly, infrequent, and spatially uneven household surveys and censuses, particularly in data-scarce regions. To address this gap, this study develops a scalable remote-sensing framework for estimating area-level presence of piped water and sewage systems at 2.56 km spatial resolution across Africa. The framework integrates Sentinel-2 imagery, enumerator-observed system-presence records from Afrobarometer enumeration areas, 30 m population data, and Vision Transformer representations learned with DINO self-supervised learning. The best-performing models achieve held-out AUROC values of 91.54\% for piped water and 93.24\% for sewage across enumeration areas. Under leave-one-region-out cross-validation, this falls to 75.5\% and 78.7\% respectively, reflecting transfer difficulty to unsampled regions. Applied across 50 African countries, population-weighted piped water estimates closely track WHO/UNICEF JMP piped water access ($R^2 = 0.92$), while sewage estimates show meaningful agreement with the broader JMP safely managed sanitation benchmark ($R^2 = 0.72$). In countries without Afrobarometer survey coverage, the model achieves population-weighted mean absolute errors of 9.5\% for piped water and 10.7\% for sewage. A Nigeria application across 767 Local Government Areas shows how our framework's fine-scale predictions reveal substantial subnational inequality, with the largest populations living where no piped water system is present reaching 1.187 million, and no sewage system 1.577 million. These findings show that DINO-based self-supervised learning using freely available satellite imagery can complement traditional household surveys, supporting SDG 6 monitoring, infrastructure planning, and environmental equity assessment.

\medskip
\noindent\textbf{Keywords:} sustainable development goals; piped water; sewage system; satellite remote sensing; self-supervised learning
\end{abstract}


\section{Introduction}
Access to safely managed drinking water and sanitation services is a fundamental prerequisite for human health and well-being. Achieving universal access to these services—the central objective of United Nations Sustainable Development Goal (SDG) 6—remains one of the most pressing and persistent development challenges of the twenty-first century. As of 2024, approximately 2.1 billion people worldwide lack access to safely managed drinking water, and 3.4 billion lack safely managed sanitation \citep{WHOJMP2025}. Sub-Saharan Africa bears a disproportionate share of this burden: only 32\% of the region's population used safely managed drinking water in 2024, compared with a global average of 74\% \citep{WHOJMP2025}. At current rates of progress, the world must accelerate efforts six-fold for drinking water and eight-fold for sanitation to meet the 2030 targets set by SDG Indicators 6.1.1 and 6.2.1 \citep{WHOJMP2025, UNWater2023}. The scale of the shortfall is compounded by a persistent financing gap: the annual investment deficit for sustainable development in developing countries now stands at approximately \$4 trillion, of which water and sanitation alone accounts for an estimated \$500 billion per year \citep{UNCTAD2023}. In this context, the ability to accurately identify where access is most deficient -- and to do so frequently, at fine spatial scales, and at low cost -- is not merely a technical convenience but a prerequisite for effective resource allocation.


Yet the data systems on which SDG 6 monitoring currently depends are poorly suited to this task. Progress toward SDG Targets 6.1 and 6.2 is tracked primarily through the WHO/UNICEF Joint Monitoring Programme (JMP), which compiles national estimates from household surveys and censuses conducted by national statistical offices \citep{JMP2024}. These surveys, principally the Demographic and Health Surveys (DHS), Multiple Indicator Cluster Surveys (MICS), and national censuses, are logistically demanding and expensive, costing an estimated \$1.3--5 million per country per round \citep{Kilic2017}.
In sub-Saharan Africa, implementation costs are particularly steep, averaging at least \$300 per surveyed household -- roughly five times the cost in other regions \citep{Kilic2017}. As a result, most countries conduct comprehensive water, sanitation and hygiene (WASH) surveys only once every three to six years, with some experiencing gaps of up to a decade between rounds. By the time survey data are collected, processed, and incorporated into JMP estimates (a pipeline that typically introduces a lag of two to four years) the resulting picture may no longer reflect conditions on the ground \citep{Fuller2016}. The JMP itself publishes updated country estimates only biennially, most recently in August 2025 \citep{WHOJMP2025}. This combination of high cost, low frequency, and long latency means that the SDG monitoring system operates with a temporal resolution fundamentally mismatched to the pace at which water and sanitation access evolves.


Equally important, the data most widely used for progress tracking are predominantly reported at the national level. Although the JMP publishes subnational disaggregations, these are generally resolved to relatively broad administrative regions and vary across countries in their geographic coverage, temporal frequency, and indicator availability, limiting their usefulness at the local scales at which infrastructure investments are planned. National and regional averages can therefore conceal wide disparities in access between urban and rural areas, between administrative regions, and even between neighboring communities \citep{UNDESA2024}. For instance, urban piped water access rates in many African countries exceed 70\%, while rural rates in the same countries fall below 15\%. Without consistently available fine-scale estimates, policymakers cannot identify the specific localities where investment is most urgently needed, and the ``SDGs for all'' vision of reaching the most remote communities remains aspirational rather than actionable \citep{IISD2021}. This information gap is particularly severe in sub-Saharan Africa, where at least 75\% of countries lack data on one or more SDG 6 indicators \citep{Dinka2024}, and only 33.5\% of SDG-related data is produced by national statistical systems \citep{UNECA2024}. Bridging this gap through traditional survey expansion is unlikely to be feasible: the World Bank has estimated that conducting the 390 surveys needed across 78 low-income countries by 2030 would require nearly \$1 billion in direct costs, with three-quarters of the burden falling on sub-Saharan Africa \citep{Kilic2017}. Alternative monitoring approaches that can deliver frequent, spatially granular, and cost-effective estimates are therefore essential.


Previous studies have sought to address these spatial data gaps through geospatial modeling. The Local Burden of Disease WaSH Collaborators used data from 600 household surveys, censuses, and other sources with Bayesian geostatistical models to estimate drinking-water and sanitation facility use on approximately 5 × 5 km surfaces across more than 88 low- and middle-income countries from 2000 to 2017, subsequently aggregating the estimates to first- and second-level administrative units \citep{Deshpande2020}. More recently, \citet{Greenwood2024} combined household-level estimates from Multiple Indicator Cluster Surveys with globally available human and environmental geospatial covariates, including Earth-observation-derived variables, in a random-forest framework. They produced subnational estimates of safely managed drinking-water service use and its four constituent service criteria across 135 low- and middle-income countries, including many countries without direct training-survey observations. These efforts demonstrate the value of survey disaggregation and geospatial modeling for identifying within-country disparities and filling monitoring gaps.


Freely available satellite imagery, combined with recent advances in deep learning, offers a promising pathway to address this monitoring deficit. The Copernicus Sentinel-2 constellation mission provides open-access multispectral imagery at 10\,m spatial resolution with a combined revisit interval as short as 2--3 days at the equator \citep{Li2020revisit, Gorelick2017} -- a temporal frequency that is orders of magnitude higher than any household survey program, at effectively zero marginal cost for data acquisition. A growing body of work has demonstrated that deep learning applied to such imagery can predict development indicators across Africa with accuracy comparable to ground surveys. \citet{Jean2016} showed that convolutional neural networks (CNNs) trained on daytime satellite imagery could explain up to 75\% of variance in local economic outcomes across five African countries; \citet{Yeh2020} scaled this approach to approximately 20,000 African villages. More directly related to infrastructure access, \citet{Oshri2018} applied CNNs to Afrobarometer-derived infrastructure labels, achieving area under the receiver operating characteristic curve (AUROC) values of 74\% for piped water and 86\% for sewerage. More broadly, \citet{Burke2021} established that satellite-based approaches can complement and extend ground-based monitoring by filling temporal and spatial gaps between survey rounds.

More recently, self-supervised learning (SSL) methods have proven particularly well-suited to remote sensing applications, where labeled data are scarce but vast archives of unlabeled imagery are readily available. SSL frameworks such as DINO \citep{Caron2021} learn transferable visual representations through self-distillation without requiring task-specific annotations, enabling robust feature extraction that generalizes across diverse geographic and environmental contexts. This paradigm has been successfully applied to a range of Earth observation tasks, including land-cover mapping \citep{Bourcier2022}, semantic segmentation \citep{Li2022}, and estimation of ground-level air quality \citep{Jiang2022}. Large-scale benchmarking efforts such as SSL4EO-S12 \citep{Wang2023ssl4eo} and GeoNet \citep{lahrichi2025self} have shown that SSL pretraining on Sentinel imagery achieves performance competitive with or exceeding supervised baselines on downstream classification tasks. Several foundation models for Earth observation have also recently emerged, including Prithvi-EO-2.0 \citep{Szwarcman2025}, Galileo \citep{Tseng2025}, CROMA \citep{fuller2023croma}, and DINOv3-sat \citep{Simeoni2025}, further demonstrating the scalability of self-supervised approaches for geospatial analysis.


Despite these advances, the application of satellite remote sensing and deep learning in water research has largely focused on natural water bodies, water quality retrieval, hydrological processes, and ecosystem monitoring \citep{Sun2024, Xiong2022, Wang2020, Hakimdavar2020, Sagan2020}. While these contributions are valuable for SDG 6 targets related to water-related ecosystems and ambient water quality (Targets 6.3 and 6.6), they do not address the infrastructure access indicators at the heart of Targets 6.1 and 6.2 -- namely, the proportion of the population with access to safely managed drinking water and sanitation services. Although a growing literature maps water, sanitation, and hygiene access subnationally from geospatial covariates and household surveys, comparatively little work has combined self-supervised representation learning of satellite imagery at continental scale with external validation of the resulting population-weighted estimates against official JMP statistics.

To address these gaps, this study develops a scalable and low-cost remote-sensing framework for estimating spatially disaggregated presence of piped water and sewage systems across Africa. In this context, piped water and sewage system presence provides infrastructure-related measures that can support fine-scale monitoring of drinking water and sanitation services under SDG 6. The framework integrates multispectral imagery from Sentinel-2, Afrobarometer survey-derived labels of area-level presence for piped water and sewage systems from Rounds 7--9 (2019--2023) \citep{AfrobarometerR7_2019, AfrobarometerR8_2022, AfrobarometerR9_2023}, and high-resolution (${\sim}30 \times 30$\,m) gridded population density data from Meta's Data for Good initiative \citep{MetaCIESIN2022}. Representations are learned using a Vision Transformer (ViT-Base) backbone trained with the DINO self-supervised framework on over 186,701 unlabeled satellite image patches, and downstream classification is performed with a lightweight $k$-nearest-neighbor (k-NN) classifier. The framework generates localized predictions at the scale of individual satellite image patches (${\sim}6$\,km$^2$), enabling identification of subnational variation in system presence not captured by conventional national-level reporting. When combined with gridded population data, these predictions yield population-weighted estimates that can serve as a spatial proxy for SDG Indicators 6.1.1 and 6.2.1. This study provides an actionable framework for SDG monitoring in data-scarce settings, with a focus on Africa, where conventional survey-based data coverage remains limited. The approach is readily transferable to other geographic contexts and potentially to other infrastructure-related SDG indicators.

\section{Methods}

\subsection{Workflow overview}

\begin{figure}[htbp]
  \centering
  \includegraphics[width=\linewidth]{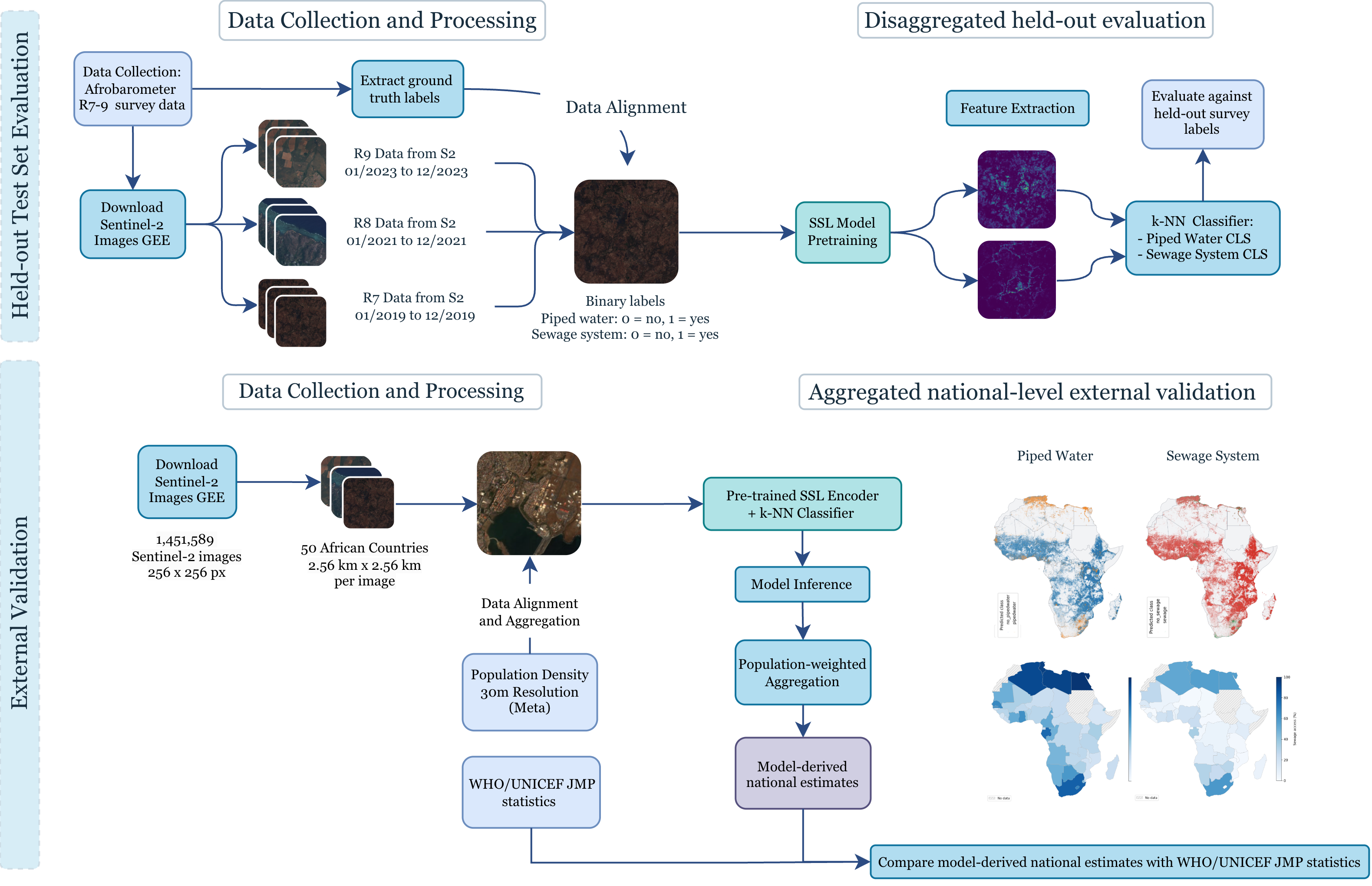}
\caption{Overview of the study workflow. The held-out evaluation pipeline constructs binary enumeration-area-level system-presence labels from Afrobarometer responses, aligns them with Sentinel-2 image patches from the corresponding survey years, extracts image features using pretrained self-supervised encoders, and evaluates $k$-NN predictions against held-out survey labels. The external-validation pipeline applies the selected model configuration across African countries, combines patch-level predictions with high-resolution population data to produce population-weighted national estimates, and compares these estimates with WHO/UNICEF JMP statistics.}
  \label{fig:workflow}
\end{figure}

As shown in Figure \ref{fig:workflow}, the analysis comprises two complementary components: disaggregated held-out test-set evaluation and aggregated national-level external validation. In the held-out evaluation pipeline, Afrobarometer survey responses are used to construct binary labels of enumeration-area-level presence of a piped water system and a sewage system, recorded by field supervisors (EA\_SVC\_B, EA\_SVC\_C), which are then aligned with Sentinel-2 image patches extracted around georeferenced survey locations \citep{AfrobarometerR7_2019, AfrobarometerR8_2022, AfrobarometerR9_2023}. These image patches are then passed through pretrained self-supervised and foundation-model encoders to generate feature embeddings, which are classified using a lightweight $k$-nearest-neighbor ($k$-NN) classifier and evaluated against held-out survey locations.

In the external-validation pipeline, the best-performing model configuration is applied to Sentinel-2 image patches across African countries to generate spatially disaggregated predictions of piped water and sewage system presence. These predictions are then combined with Meta’s High Resolution Population Density Maps to derive population-weighted national estimates, which are compared with official WHO/UNICEF Joint Monitoring Programme (JMP) statistics corresponding to SDG Indicators 6.1.1 and 6.2.1 \citep{JMP2024, MetaCIESIN2022}.

\subsection{Data sources and processing}

\label{sec:data_sources_processing}

This study draws on four primary data sources: Afrobarometer survey data, Sentinel-2 satellite imagery, Meta’s High Resolution Population Density Maps, and official national statistics from the JMP.

\begin{figure}[htbp]
\centering

\begin{subfigure}{\linewidth}
  \centering
  \includegraphics[width=0.7\linewidth]{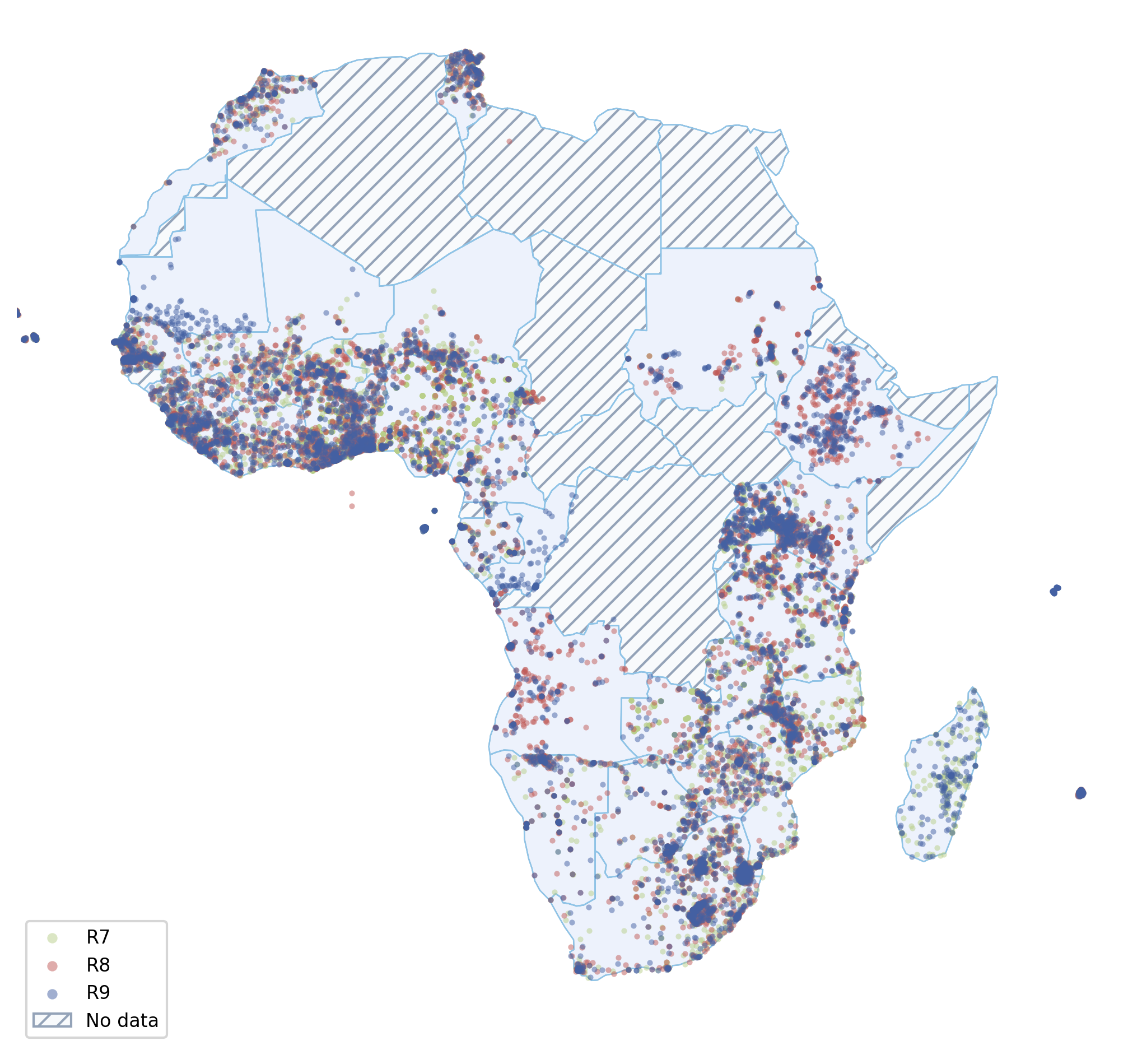}
  \caption{Spatial distribution of Afrobarometer survey locations from Rounds 7--9 (2019--2023).}
  \label{fig:survey_map}
\end{subfigure}

\vspace{0.5em}

\begin{subfigure}{0.48\linewidth}
  \centering
  \includegraphics[width=\linewidth]{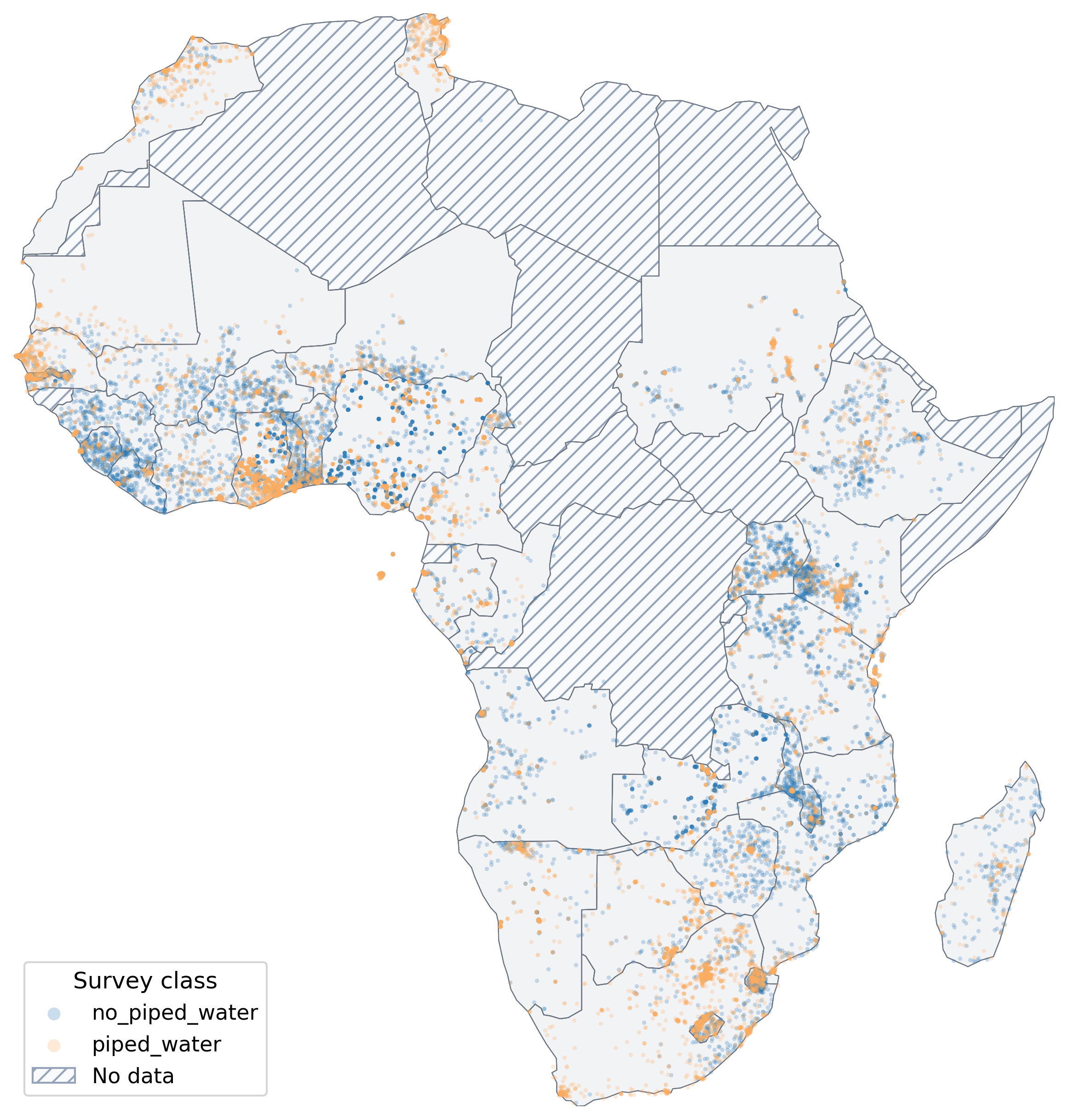}
  \caption{Survey-derived piped water system presence}
  \label{fig:survey_pw_map}
\end{subfigure}
\hfill
\begin{subfigure}{0.48\linewidth}
  \centering
  \includegraphics[width=\linewidth]{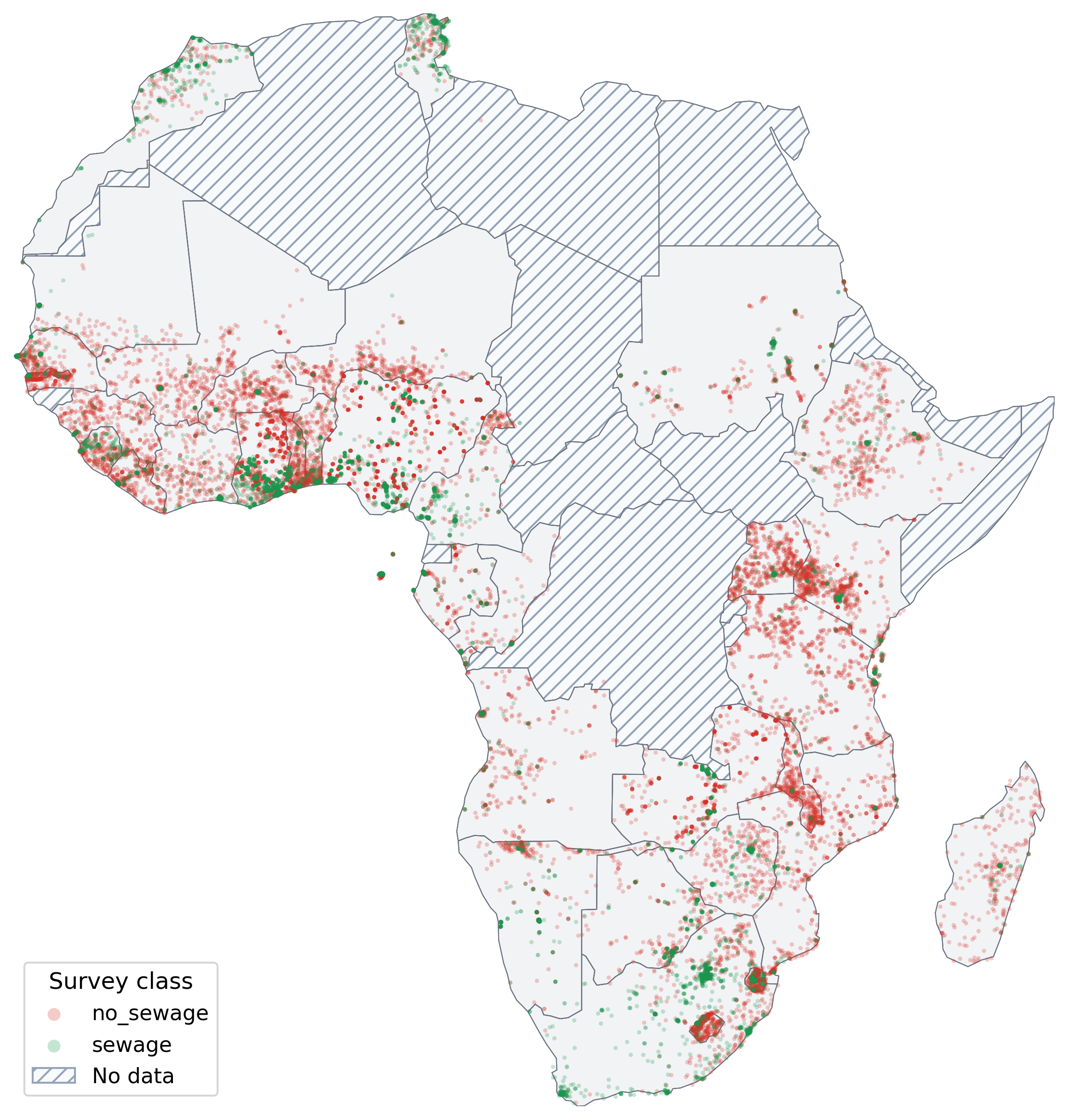}
  \caption{Survey-derived sewage system presence}
  \label{fig:survey_sw_map}
\end{subfigure}

\caption{Afrobarometer survey coverage and survey-derived system-presence labels used for model training and evaluation. Panel (a) shows the spatial distribution of georeferenced Afrobarometer survey locations from Rounds 7--9 (2019--2023). Panels (b) and (c) show binary survey-derived labels for piped water and sewage system presence, respectively. Hatched countries indicate areas without Afrobarometer survey coverage in the study sample.}
\label{fig:combined_survey_maps}

\end{figure}

Afrobarometer survey data from Rounds 7 (2019), 8 (2021), and 9 (2023) were used to construct survey-derived binary labels of area-level presence of piped water and sewage systems across 40 of Africa’s 54 countries \citep{AfrobarometerR7_2019, AfrobarometerR8_2022, AfrobarometerR9_2023}. These labels derive from the Afrobarometer enumerator-completed enumeration-area (EA) observation of whether piped water and sewage systems were present in the sampled area (variables \texttt{EA\_SVC\_B} and \texttt{EA\_SVC\_C}); we aggregated to the EA GPS point and assigned a positive label where the system was recorded as present. They therefore represent area-level system presence rather than household-level service use, and are best interpreted as a spatial proxy for -- rather than a direct measurement of -- SDG Indicators 6.1.1 and 6.2.1. Across these rounds, the dataset comprised 45,823, 48,084, and 53,444 georeferenced interviews, corresponding to 7,704, 5,151, and 5,517 unique survey locations, respectively.
A survey location here is an enumeration area, the primary sampling unit at which Afrobarometer records GPS coordinates; each location contributes multiple individual interviews (approximately six to eight on average), and all interviews within a location share a single coordinate pair and therefore a single Sentinel-2 patch. This is why the analysis splits and the cross-validation folds are constructed over locations rather than interviews. Afrobarometer does not apply random displacement to released EA coordinates, unlike the DHS, so no positional offset is introduced between the label and the image patch. The spatial distribution of survey locations and the corresponding piped water and sewage system presence labels are shown in Figure~\ref{fig:combined_survey_maps}.

Afrobarometer uses a clustered, stratified, multistage probability sample designed to be nationally representative of the voting-age population, with primary sampling units selected with probability proportional to population size and stratified by region and by urban--rural status; sample sizes are typically 1,200--2,400 respondents per country per round, yielding the enumeration-area counts reported above. We used Afrobarometer rather than DHS or MICS despite the closer alignment of the latter with JMP service ladders for three reasons. First, Rounds 7--9 provide near-contemporaneous coverage of 40 countries on a common instrument, whereas DHS and MICS rounds are fielded on country-specific schedules that would introduce substantial temporal heterogeneity across a continental training set. Second, the enumerator-completed EA observation used here records infrastructure presence for the sampled area as a whole, which matches the $\sim$2.56\, km image patch more directly than a household-level service-use response would. Third, Afrobarometer releases georeferenced EA coordinates under terms that permit the image-alignment step central to this framework. The trade-off is that our labels measure area-level infrastructure presence rather than directly measuring the safely managed service use captured by Indicators 6.1.1 and 6.2.1.

Sentinel-2 imagery was retrieved through Google Earth Engine (GEE) for these locations using monthly observations from the calendar year corresponding to each survey round, aligning satellite data with survey timing while reducing seasonal mismatch. Images were then cropped into 256 $\times$ 256-pixel patches centered on survey locations, subject to a maximum cloud-cover threshold of 10\% for quality control. Sentinel-2 provides multispectral observations at 10 m/pixel spatial resolution for the red, green, and blue (RGB) bands used in this study, enabling representation of fine-scale built-environment features relevant to infrastructure presence \citep{Gorelick2017}. The analysis uses only the RGB bands for four reasons. First, the visual cues that distinguish presence from absence locations in this setting are primarily spatial and morphological (settlement density, road-network structure, and building footprints) rather than the spectral land-cover contrasts for which the red-edge, NIR, and SWIR bands are most informative, and the RGB bands are delivered at Sentinel-2's finest (10\,m) native resolution. Second, a three-channel input is directly compatible with the RGB-pretrained DINO/ViT encoders compared here, avoiding modification of the patch-embedding layer. Third, RGB preserves comparability with prior satellite-based development-mapping work over Africa \citep{Jean2016, Yeh2020, Oshri2018}. Fourth, an RGB-only pipeline is more portable and operationally lightweight: it can be applied to any source of visible-band imagery, including aerial, drone, and commercial basemap products, and carries a smaller data, storage, and compute footprint, lowering the barrier to adoption by national statistical offices and other teams monitoring SDG~6 in low-resource settings. We did not re-export and re-pretrain on the full multispectral stack; whether the additional red-edge, NIR, and SWIR bands would improve discrimination is an open question that would require multi-channel self-supervised pretraining, which we leave to future work.

Population-weighted estimates were derived for the 50 African countries present in Meta's High Resolution Population Density Maps \citep{MetaCIESIN2022}, extending the model inference domain beyond the 40 countries represented in the Afrobarometer survey data. These maps provide gridded population estimates at ~30 × 30 m resolution, produced by combining census-derived demographic data with machine-learning-based building footprints extracted from satellite imagery. For each image patch, population counts were aggregated by summing all grid cells whose centroids fell within the patch extent; the model’s binary access predictions (0/1) were then applied to these counts to produce population-weighted estimates at regional and national scales.

Official national-level statistics were obtained from the WHO/UNICEF JMP and used for external validation of model-derived national estimates \citep{JMP2024}. For each country, we used the most recent JMP national estimate available in the JMP 2024 country files (reference year 2022). For piped water, we compared model-derived estimates with the JMP piped drinking water access rate, which most closely corresponds to the Afrobarometer piped-water variable. For sewage system access, JMP does not provide a directly equivalent sewage-system indicator; we therefore used the safely managed sanitation access rate as the nearest available benchmark, because the model target is area-level system presence rather than household sanitation use. Afrobarometer contains no household-level sewer-connection item, so an area-level label is the only construct the survey supports for this service. This sanitation comparison should be interpreted as an approximate external validation rather than a strict one-to-one indicator match, because safely managed sanitation includes a broader set of facility types, including sewer connections, septic systems, and latrines, than the Afrobarometer-derived sewage-system access variable. The external-validation set was defined by countries with both model-derived estimates and corresponding JMP national statistics. This set includes countries represented in the Afrobarometer survey data as well as countries without Afrobarometer survey coverage, allowing us to assess model performance within the survey-country domain and in settings where no Afrobarometer training or evaluation data were available. The satellite imagery underlying the large-scale inference pools observations from the three Afrobarometer rounds (2019, 2021, and 2023), so the model-derived national estimates correspond to an imagery window centered near 2021, whereas the JMP benchmark refers to 2022. This implies an offset of roughly one year from the imagery midpoint, and at most about three years relative to the earliest (2019) round. Because the JMP national figures are smoothed annual estimates fitted by regression through the underlying surveys and censuses, and are themselves revised only biennially, national coverage changes only incrementally over a window of this length; we therefore treat the residual temporal offset as a minor source of noise in the country-level comparison rather than a systematic bias, while noting that it contributes to the scatter in Figure~\ref{fig:scatter}.

\subsection{Model comparison and representation learning}

To evaluate self-supervised representation learning for this task, we compared pretrained Vision Transformer (ViT) feature extractors for satellite imagery. In the main text, we focus on the DINO family (DINO, DINOv2, and DINOv3) while results for additional Earth observation foundation models, including Galileo and Prithvi-EO-2.0, are reported in Supplementary Material \citep{Caron2021, Oquab2023, Simeoni2025, Tseng2025, Szwarcman2025}.


Although piped water and sewage infrastructure are not always directly visible in satellite imagery, surrounding land-use and built-environment characteristics can systematically co-occur with infrastructure presence. As illustrated in Figure~\ref{fig:piped_water_examples}, image patches from areas with and without piped water presence show visible differences in settlement density, road network structure, agricultural patterns, and proximity to developed areas. These spatial patterns provide indirect signals from which self-supervised visual representations may learn features relevant to piped water and sewage system presence.

DINO learns representations through self-distillation without labels using a student and a teacher network with the same architecture \citep{Caron2021}. The student processes both global and local crops, whereas the teacher processes only global crops, and the student is trained to match the teacher's output distribution across views (Figure~\ref{fig:dino_framework}). The objective is a cross-entropy loss,
\[
\mathcal{L} = - \sum p_t(x)\log p_s(x),
\]
where $p_t(x)$ and $p_s(x)$ denote the teacher and student output distributions, respectively. The teacher is not directly optimized by the loss; instead, its weights are updated as an exponential moving average of the student weights, producing a more stable encoder for representation learning \citep{Caron2021}. After pretraining, we use the teacher encoder to extract image embeddings for downstream $k$-NN classification.

Three features of DINO suit this application. First, we have far more
imagery than labels: about $10^4$ georeferenced survey locations, against more than
1.4 million unlabelled Sentinel-2 patches across the inference domain. Self-supervised
pretraining uses all of the imagery to learn settlement structure, so the labels are
needed only for the final classifier. Second, the multi-crop scheme trains the student
to match the teacher across both local and global views of a scene. This produces
representations that link small built-environment details to their wider surroundings,
which is what matters here, since the signal we rely on is the shape and arrangement
of settlement rather than its spectral properties (Figure~S1). Third, DINO
representations classify well with nearest neighbours and without fine-tuning
\citep{Caron2021}. Hence, the downstream step can be a simple $k$-NN classifier,
which fits very few parameters to the survey labels.

We compared three DINO-family models. DINO and DINOv2 were pretrained in this study on over 1.4 million unlabeled Sentinel-2 image patches of size $256 \times 256$ pixels. DINOv2 extends the original DINO framework through curated large-scale pretraining and combined image-level and patch-level self-supervision, while DINOv3 further improves representation quality at scale through refinements such as Gram anchoring \citep{Oquab2023, Simeoni2025}. DINOv3 and the additional foundation models were evaluated using publicly available pretrained weights, since these models were originally trained at scales beyond the data and compute available in this study. For each geolocated survey location, each model produced an image embedding summarizing the surrounding built environment, enabling systematic comparison of how well different pretrained representations capture infrastructure-related patterns across diverse African contexts.

\subsection{Held-out test-set evaluation}

To reduce spatial data leakage, we split the data at the survey-location level rather than the individual interview level, assigning locations to training, validation, and test sets in an 80\%/10\%/10\% ratio. For each model, embeddings extracted from Sentinel-2 image patches were paired with survey-derived binary labels for piped water and sewage system presence. Classification was performed in the embedding space using a lightweight $k$-nearest-neighbor ($k$-NN) classifier, with the value of $k$ selected separately for each model--task pair using validation-set performance.

Held-out performance was evaluated on survey locations in the test set using accuracy, macro-averaged recall, macro-averaged F1-score -- the harmonic mean of precision and recall, computed separately for the presence and absence classes and then averaged unweighted, so that both classes contribute equally despite their unequal prevalence- and area under the receiver operating characteristic curve (AUROC). These metrics were used to compare model configurations, and the best-performing configuration was then used for large-scale inference and external validation.

Because a random location-level split can still place spatially adjacent locations in different splits, and the framework's central claim is transfer to unsampled regions, we additionally report leave-one-region-out cross-validation over the five UN African subregions. The definitions of the five subregions and the corresponding cross-validation folds are provided in Supplementary Material~\ref{sec:appendix_robustness} and Figure~\ref{fig:region_folds_cv}

\subsection{External validation}

External validation against independent official statistics was used to assess whether patch-level satellite predictions, after large-scale deployment and population-weighted aggregation, could support SDG 6 monitoring. This step evaluates the policy-facing output of the framework, rather than only its classification performance at Afrobarometer survey locations. Using the selected model configuration, we generated spatially disaggregated predictions of piped water and sewage system presence across the African inference domain. These predictions were aggregated to population-weighted national estimates and compared with the corresponding WHO/UNICEF JMP national statistics, following the data-processing and benchmark definitions described in Section~\ref{sec:data_sources_processing}.

\subsection{Training Setup}

DINO and DINOv2 were pretrained on 8 NVIDIA A40 GPUs using distributed training. For DINO, we used a ViT-Base backbone with patch size 8, a batch size of 16 per GPU, and 300 training epochs. For DINOv2, we used a ViT-Large backbone with patch size 8, a batch size of 16 per GPU, and 200 training epochs with 20 warmup epochs. Optimization was performed with AdamW. The initial learning rate was set to $3\times 10^{-4}$ for DINO, while the remaining DINOv2 hyperparameters followed the specified training configuration and official defaults of the DINOv2 codebase.

For held-out evaluation, embeddings were extracted from each pretrained encoder and classified using a $k$-NN classifier with cosine similarity, softmax weighting, and temperature 0.07. For the DINO and DINOv2 models pretrained in this study, we used the teacher encoder for embedding extraction. We evaluated $k \in \{5, 10, 20, 50, 100, 200\}$ for all models on an identical grid. Models evaluated using publicly available checkpoints, including DINOv3, Galileo, and Prithvi-EO-2.0, were not retrained; the released pretrained encoders were used directly under the same held-out evaluation setup. Additional implementation details are provided in Table~\ref{tab:training_details}.
\section{Results and Discussion}

\subsection{Disaggregated held-out evaluation of piped water and sewage system presence}

\subsubsection{Held-out predictive performance and model comparison}
The presence of a piped water system or sewage system in a locality is a precondition for safely managed drinking water and sanitation services, making them directly relevant to SDG 6 Targets 6.1 and 6.2 \citep{UN2015a}, though it is not equivalent to household use of those services. To assess whether pretrained satellite-image representations can support disaggregated prediction of system presence, we evaluated multiple ViT-based models on held-out survey locations using binary labels of piped water and sewage system presence.

Table~\ref{tab:accuracy} reports held-out test-set performance for each model and task, with the value of $k$ in the $k$-nearest-neighbor classifier selected using validation-set performance. Because predictions are made for 256 $\times$ 256-pixel Sentinel-2 patches, corresponding to approximately 2.56 km $\times$ 2.56 km on the ground, these metrics evaluate spatially disaggregated performance rather than performance after aggregation to regional or national units. We report accuracy, macro-averaged recall, macro-averaged F1-score, and AUROC to summarize classification performance across presence and absence locations. Overall, the best-performing models achieved strong performance for both piped water and sewage system presence prediction, indicating that pretrained satellite-image representations capture built-environment characteristics associated with system presence. At a matched neighborhood size ($k=200$), DINO and DINOv2 were statistically indistinguishable, with AUROC of 91.54\% versus 91.54\% for piped water and 93.33\% versus 93.24\% for sewage system; within the DINO family, the choice of encoder therefore had little effect on this task. DINOv3 using the released SAT-493M satellite-imagery checkpoint as off-the-shelf weights consistently achieved AUROC scores approximately 3--4 percentage points lower than those of the models pretrained in this study. Additional results for other Earth observation foundation models, including Galileo and Prithvi-EO-2.0, are reported in Table~\ref{tab:foundation_model_summary} and discussed in Supplementary Material.


These results represent a substantial improvement over prior work. For example, \citet{Oshri2018} applied CNNs to Afrobarometer-derived labels and reported AUROC values of 74\% for piped water and 86\% for sewerage presence. In comparison, the higher AUROC values achieved in this study suggest that self-supervised Vision Transformer--based representations provide stronger discriminative performance for distinguishing locations with and without infrastructure.

Across models, sewage system presence was predicted slightly more accurately than piped water presence. One possible explanation is that sewage system presence is more strongly associated with dense urban development and visible built-environment features captured by Sentinel-2 imagery, such as settlement density, road networks, and larger building footprints. In contrast, piped water presence may occur in settings where these surface-visible proxies are less distinct.

\begin{table}[htbp]
  \centering
  \caption{Held-out performance (\%) by model and task using the validation-selected value of $k$. Recall and F1 are macro-averaged across classes. DINO and DINOv2 are evaluated at the matched neighborhood size $k=200$ and differ by less than run-to-run variation, so neither is marked as best; DINOv3 uses publicly available weights at its validation-selected $k$. Under ``Pretrain data'', \emph{ours} denotes self-supervised pretraining performed in this study on over 1.4 million unlabeled Sentinel-2 patches from the African inference domain, and \emph{sat (orig.)} denotes the released SAT-493M satellite-imagery checkpoint distributed with DINOv3.}
  \label{tab:accuracy}
  \small
  \resizebox{\textwidth}{!}{%
  \begin{tabular}{llcccrrrr}
    \toprule
    \textbf{Task} & \textbf{Model} & \textbf{Backbone} & \textbf{Pretrain data} & \textbf{Best $k$} & \textbf{Accuracy} & \textbf{Recall} & \textbf{F1} & \textbf{AUROC} \\
    \midrule
    \multirow{3}{*}{Piped Water}
      & DINOv3 & ViT-L/16 & sat (orig.) & 10 & 81.43 & 81.22 & 81.19 & 88.25 \\
      & DINOv2 & ViT-L/8 & ours & 200 & \textbf{84.13} & \textbf{83.76} & \textbf{83.85} & \textbf{91.54} \\
      & DINO & ViT-B/8 & ours & 200 & 83.88 & 83.47 & 83.58 & 91.54 \\
    \midrule
    \multirow{3}{*}{Sewage System}
      & DINOv3 & ViT-L/16 & sat (orig.) & 10 & 84.65 & 81.06 & 81.92 & 89.17 \\
      & DINOv2 & ViT-L/8 & ours & 200 & 87.17 & 84.29 & 85.00 & 93.24 \\
      & DINO & ViT-B/8 & ours & 200 & \textbf{87.23} & \textbf{84.54} & \textbf{85.13} & \textbf{93.33} \\
    \bottomrule
  \end{tabular}%
  }
\end{table}

\subsubsection{Generalization to unsampled regions}

Because a central application of the framework is estimating system presence in regions and countries without survey coverage, we assessed the geographic transferability of the learned representations. In addition to the original 80\%/10\%/10\% train/validation/test split and random 5-fold cross-validation over survey locations, we conducted leave-one-region-out cross-validation across the five UN African subregions---Northern, Western, Middle, Eastern, and Southern Africa (fold assignments in Figure~\ref{fig:region_folds_cv}; performance comparisons in Table~\ref{tab:split_schemes}). In each fold, all survey locations within one subregion were excluded from training and reserved for evaluation. This design provides a substantially more demanding test of geographic transfer than a random location split, although some training locations may remain geographically close to held-out locations across subregional boundaries. Under this stringent evaluation, the DINOv2 embeddings achieved mean held-out AUROC values of 75.5\% for piped water and 78.7\% for sewage systems, compared with 91.3\% and 92.5\%, respectively, under random 5-fold cross-validation. Although lower than the random cross-validation results, these out-of-region AUROC values remained well above chance, suggesting that the self-supervised representations learned from large collections of unlabeled satellite imagery encode transferable visual information related to infrastructure presence. The ability to generalize across geographically distinct regions indicates that these representations capture features that are not confined to local landscape characteristics, supporting their application to piped
water and sewage systems-presence estimation in areas lacking survey observations.

\subsubsection{Substantial predictive value of satellite imagery beyond settlement type and population density}

Area-level presence of piped water and sewage systems is associated with settlement type and population density, raising the possibility that the learned representations merely detect broad patterns of urbanization rather than infrastructure-relevant characteristics. Table~\ref{tab:urban_rural_summary} shows that urban enumeration areas have substantially higher access rates than rural enumeration areas across all three survey rounds, confirming that settlement type alone provides a strong predictive signal. To quantify the additional value of satellite imagery beyond these readily available covariates, we benchmarked the image embeddings against three covariate-only baselines under random 5-fold cross-validation (Table~\ref{tab:stratified}): an urban--rural indicator, a population-density classifier, and a model combining both variables. Each baseline was fitted on the training folds and evaluated on the corresponding held-out fold using the same procedure as the image embeddings. Across the two infrastructure outcomes, the image embeddings exceeded the urban--rural baseline by 14.7--16.1 AUROC percentage points and the combined urban--rural and population-density baseline by 13.3--14.3 percentage points; the population-density-only baseline by $\sim$ 22.0 percentage points. Importantly, when urban and rural locations were evaluated separately---holding settlement type constant by design---the embeddings retained AUROC values of 85.9--88.2\%. Together, these results show that the learned representations capture infrastructure-relevant visual information that cannot be reduced to broad urbanization patterns, demonstrating the substantial incremental predictive value of satellite imagery over covariate-only approaches.

\subsection{Aggregated national-level external validation against JMP statistics}

\begin{figure}[htbp]
\centering

\begin{subfigure}{0.40\linewidth}
  \centering
  \includegraphics[width=\linewidth]{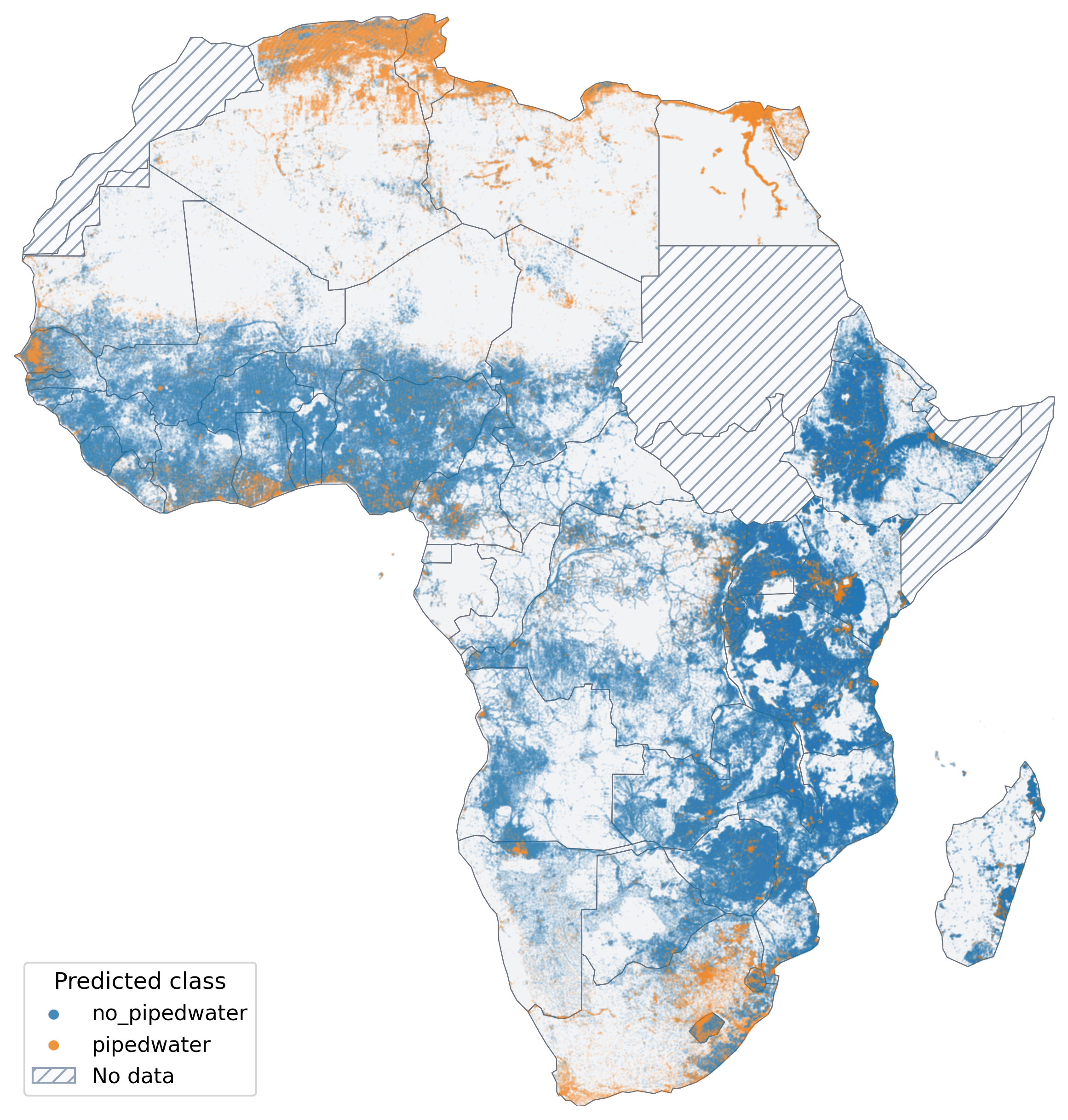}
  \caption{Disaggregated piped water results}
  \label{fig:piped_water_inference_map}
\end{subfigure}
\hspace{0.04\linewidth}
\begin{subfigure}{0.40\linewidth}
  \centering
  \includegraphics[width=\linewidth]{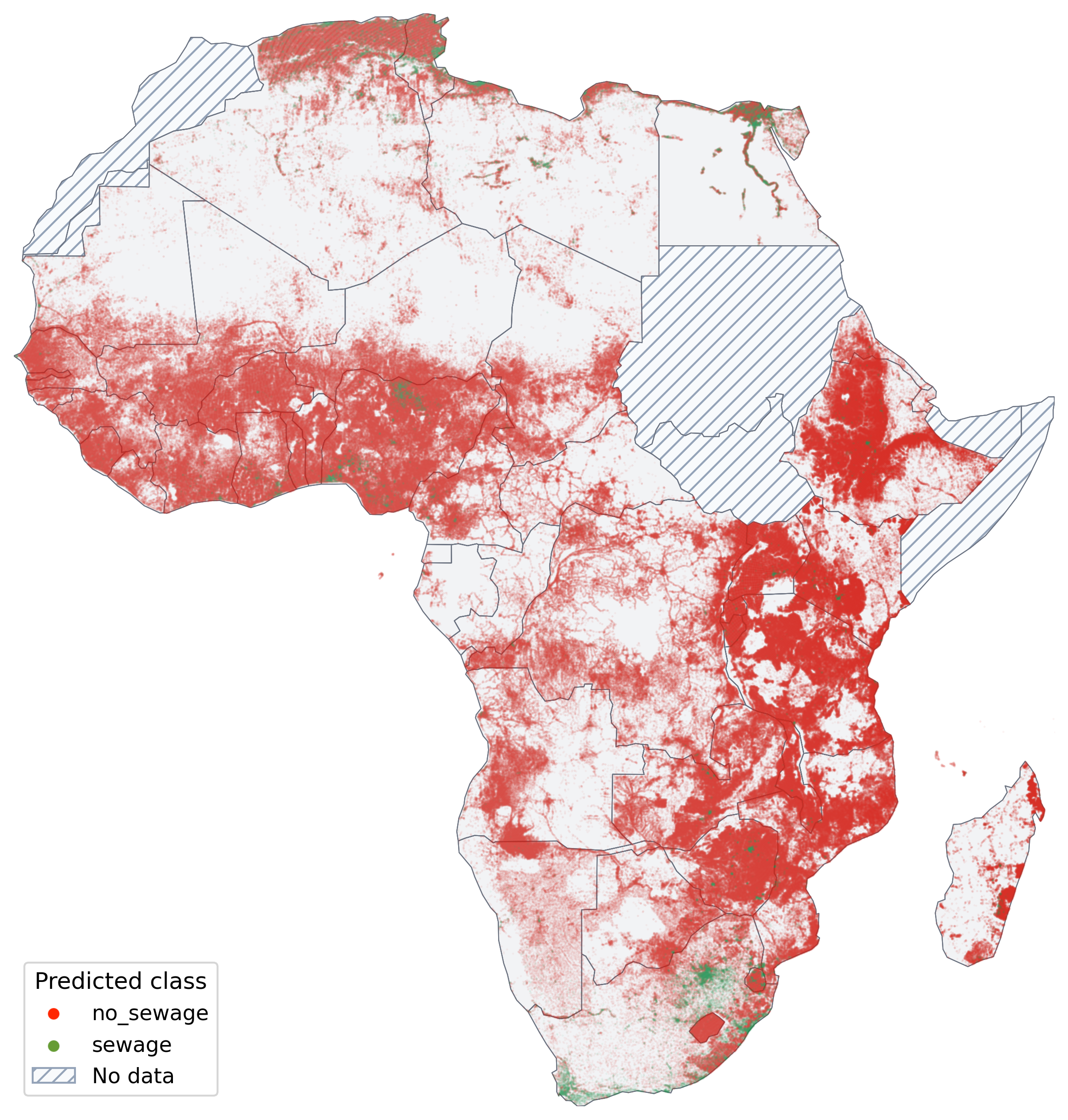}
  \caption{Disaggregated sewage system results}
  \label{fig:sewage_inference_map}
\end{subfigure}

\vspace{2em}

\begin{subfigure}{0.48\linewidth}
  \centering
  \includegraphics[width=\linewidth]{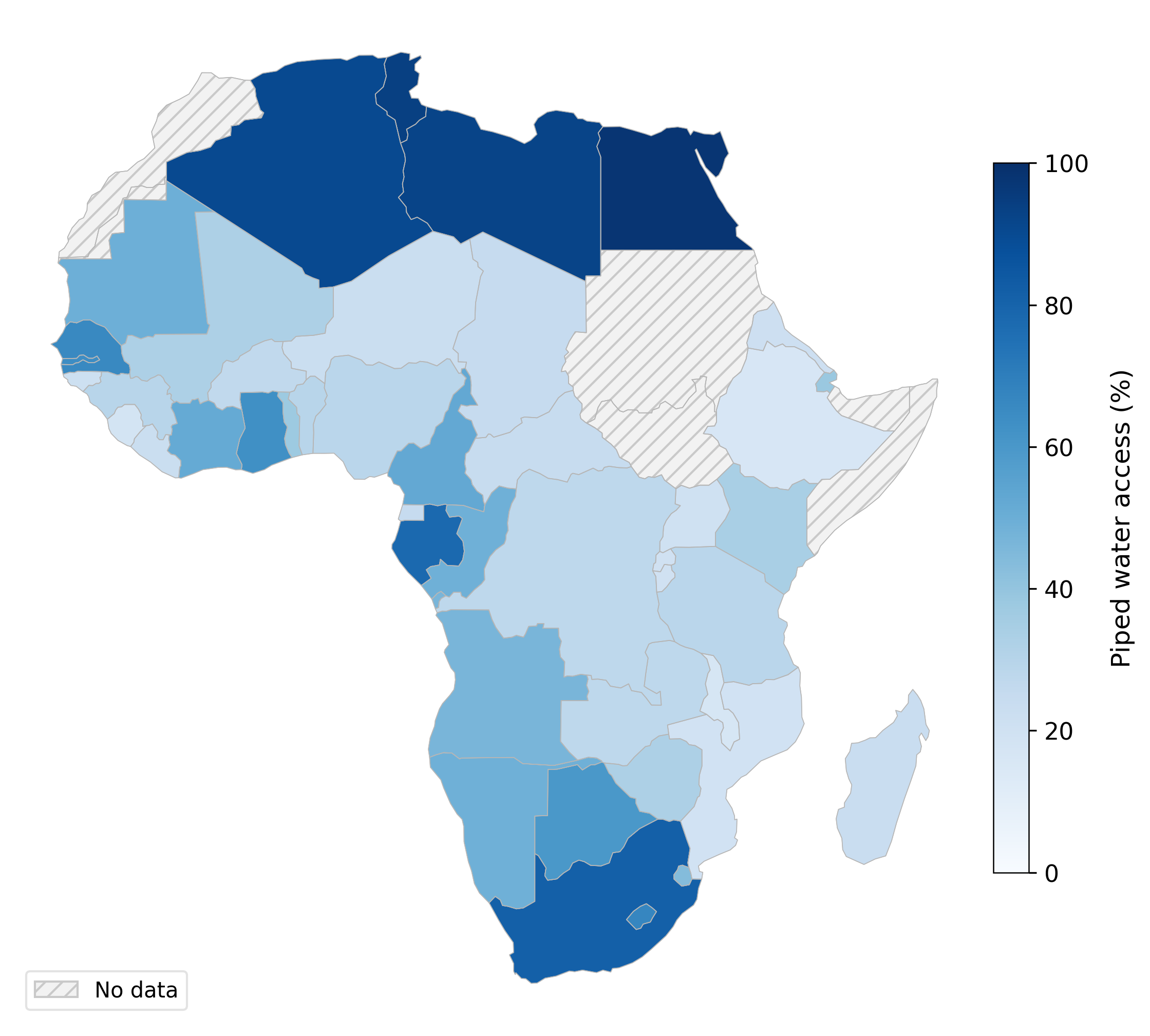}
  \caption{Population-weighted piped water presence}
  \label{fig:piped_water_map}
\end{subfigure}
\hfill
\begin{subfigure}{0.48\linewidth}
  \centering
  \includegraphics[width=\linewidth]{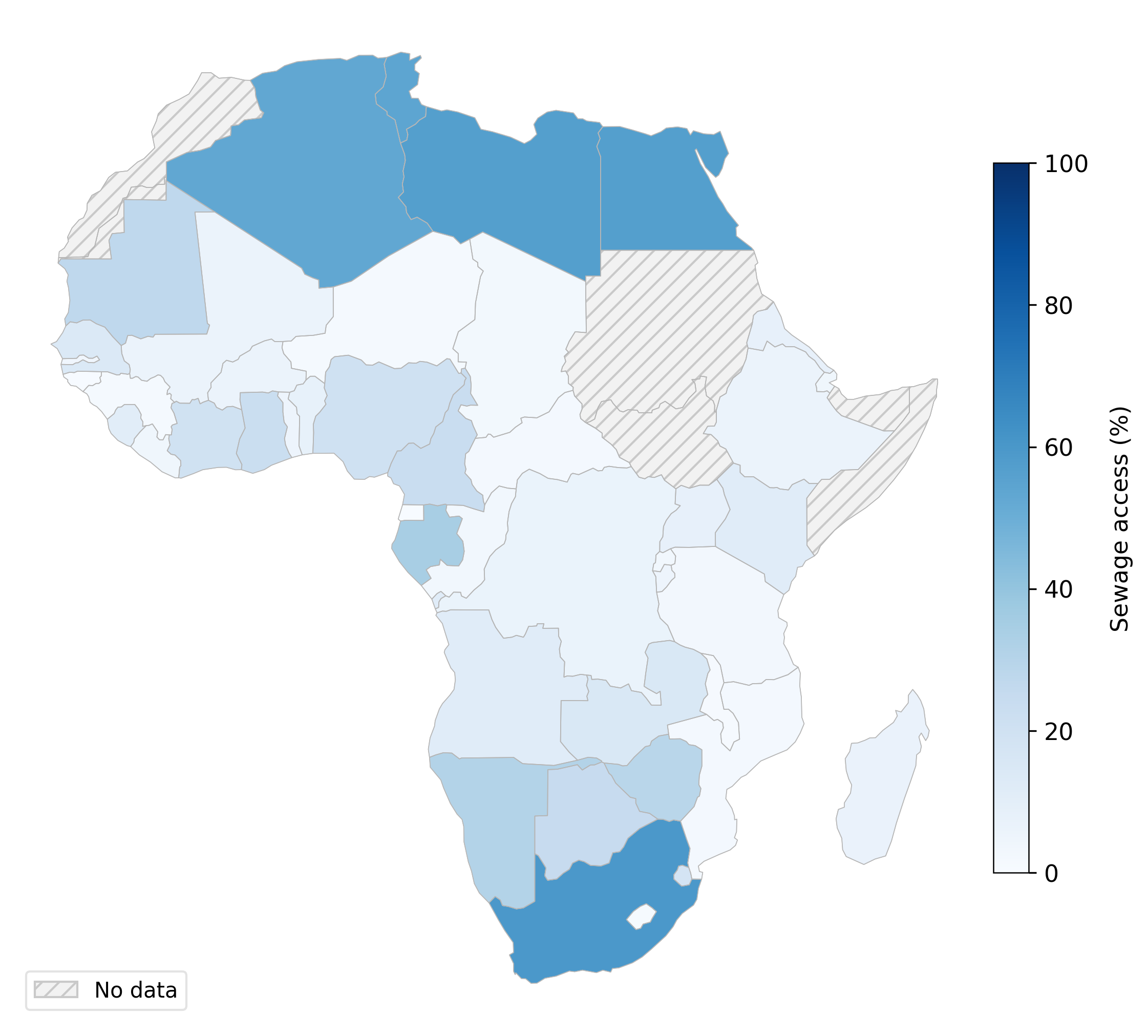}
  \caption{Population-weighted sewage system presence}
  \label{fig:sewage_map}
\end{subfigure}

\caption{Model-derived estimates of piped water and sewage system presence across Africa. Panels (a) and (b) show disaggregated patch-level predictions for piped water and sewage system presence at approximately 2.56 km spatial resolution. Grid cells with zero population are not shown, which accounts for the apparent gaps in the disaggregated maps. Panels (c) and (d) show the corresponding population-weighted national access estimates derived by combining patch-level predictions with gridded population data. Hatched countries indicate areas without population data in Meta's High Resolution Population Density Maps and are excluded from large-scale inference.}
\label{fig:access_maps}
\end{figure}

To assess whether fine-scale satellite-based predictions could be aggregated into reliable national estimates for SDG 6 monitoring, we applied the best-performing model configuration to more than 1.4 million Sentinel-2 image patches across 50 African countries, generating spatially disaggregated predictions of piped water and sewage system presence at approximately 2.56 km $\times$ 2.56 km resolution (Figures~\ref{fig:piped_water_inference_map} and \ref{fig:sewage_inference_map}). These patch-level predictions were aggregated to population-weighted national estimates using gridded population counts from Meta's High Resolution Population Density Maps (Figures~\ref{fig:piped_water_map} and \ref{fig:sewage_map}). The resulting estimates were then compared with official WHO/UNICEF JMP national statistics used as external benchmarks for SDG Indicators 6.1.1 and 6.2.1.

\begin{figure}[htbp]
  \centering
  \includegraphics[width=0.95\linewidth]{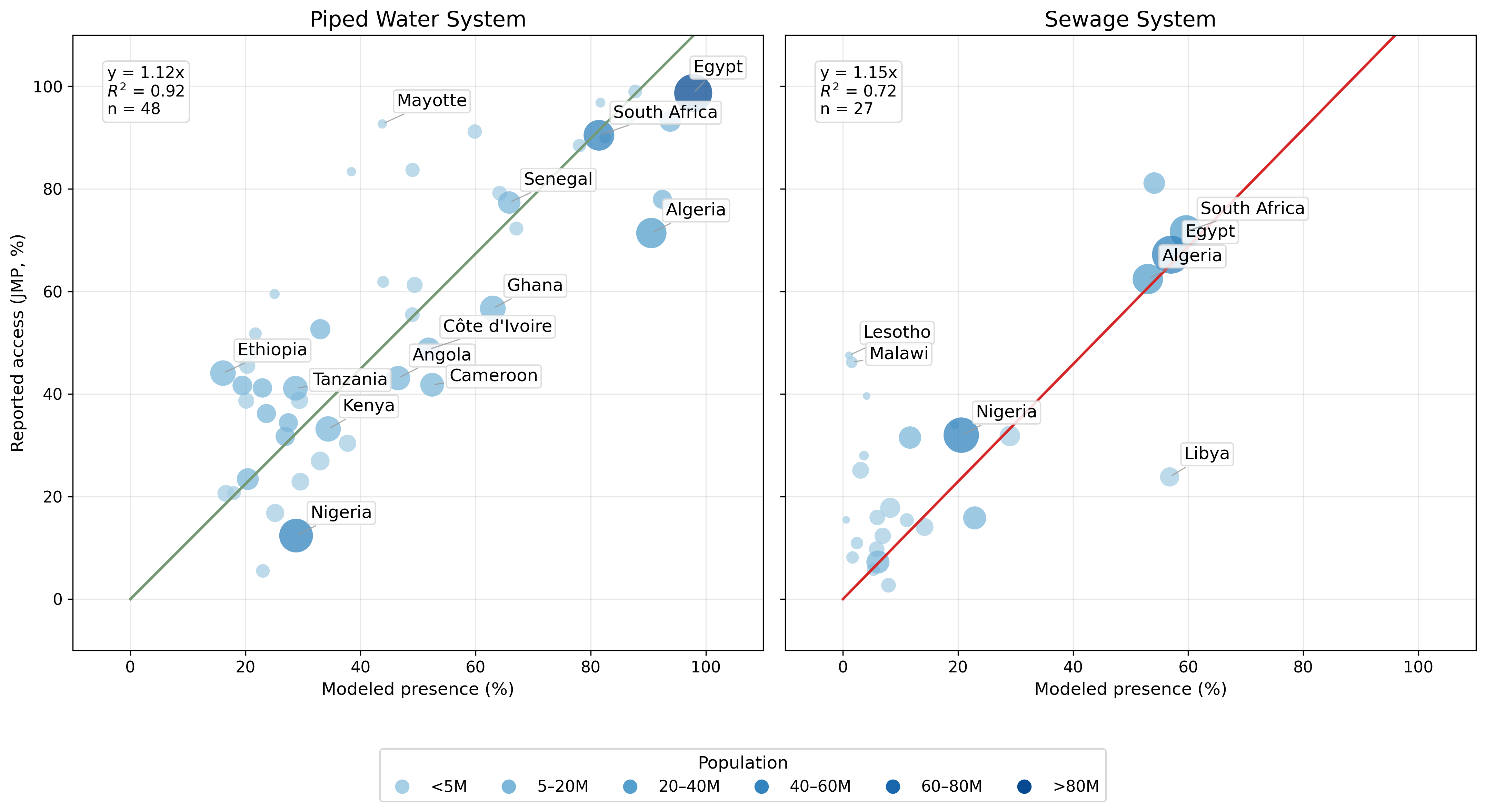}
  \caption{External validation of model-derived national access estimates against WHO/UNICEF JMP statistics. The left panel compares piped water presence estimates with JMP piped drinking water access rates, and the right panel compares sewage system presence estimates with JMP safely managed sanitation access rates. Each point represents a country; bubble size indicates total population. The x-axis reports modeled area-level system presence and the y-axis reports JMP household-use rates; the two measure related but non-identical constructs, and the comparison is an external benchmark rather than an indicator-equivalent validation.}
  \label{fig:scatter}
\end{figure}

 Figure~\ref{fig:scatter} evaluates model-derived national estimates with JMP-reported values for piped water and sewage/sanitation access. Because the model target is area-level system presence while JMP reports household use, some divergence is expected by construction. For piped water, model-derived estimates showed strong agreement with official statistics, with a fitted slope of 1.12 and $\mathrm{R}^2 = 0.92$ (Root Mean Square Error (RMSE) = 18.2 percentage points and Mean Absolute Error (MAE) = 14.3 percentage points across $n=48$ countries); for sewage, the fitted slope is 1.15 and $\mathrm{R}^2$ is 0.72 (RMSE = 19.3 and MAE = 14.4 across $n=27$). This indicates that satellite-based predictions capture meaningful national-scale variation in piped water access (SDG 6.1), although the model tends to slightly underestimate coverage in some countries relative to JMP-reported values. One possible explanation is that administratively reported service coverage may include infrastructure that is not readily distinguishable from optical satellite imagery, and that such infrastructure accounts for a larger share of reported
coverage where visible built-environment proxies are weakest. The practical
consequence is that the framework is more likely to understate than to overstate coverage in the least-served countries, which is where prioritization decisions are most consequential.

The two fits differ in how much they depend on a few high-access, high-population countries--principally Egypt and South Africa-- which sit at the top of the range and exert substantial leverage. The piped-water fit is robust to them: excluding the two leaves $\mathrm{R}^2$ essentially unchanged (0.92 to 0.92). The sanitation fit is more sensitive: excluding them lowers $\mathrm{R}^2$ from 0.72 to 0.61, so part of the sanitation agreement is carried by these leverage points. We therefore report the fitted slope alongside RMSE and MAE rather than relying on $\mathrm{R}^2$ alone. It is worth noting that we do not apply a bias correction to the estimates presented here, because the correction would be estimated from the same national statistics against which the framework is being validated and would not be identifiable in the countries lacking JMP coverage, which are precisely the setting the framework targets. Users applying these estimates for prioritization within a country with a JMP benchmark could rescale by the fitted slope; users applying them where no benchmark exists should treat the estimates as conservative.

Agreement with JMP statistics was weaker for sewage than for piped water, with $\mathrm{R}^2 = 0.72$ compared with 0.92, despite stronger performance at the spatially disaggregated level. These two observations are connected. Corresponding JMP data were available for fewer countries for sanitation ($n=27$) than for piped water ($n=48$), and that smaller sample is also what makes the fit sensitive to the leverage points identified above: Egypt and South Africa carry proportionally more weight among 27 countries than among 48. The sanitation agreement therefore rests both on fewer countries and disproportionately on a few of them, which is why we report slope, RMSE, and MAE alongside $\mathrm{R}^2$ for this indicator in particular. The fitted slope for sewage was 1.15, close to the piped water slope of 1.12, indicating proportional bias of comparable direction and magnitude relative to JMP statistics. This relatively modest difference in slope is encouraging given the benchmark mismatch: JMP sanitation indicators encompass multiple facility types, including sewer connections, septic systems, and latrines, whereas the survey-derived labels and model outputs in this study more specifically capture sewage-related infrastructure presence \citep{JMP2024}.

We note that agreement with JMP statistics is one validation criterion rather than a ground truth. Greenwood et al. (2024) report subnational estimates of safely managed drinking water use implying substantially lower coverage than JMP figures for the same indicator, illustrating that different modeling choices and indicator definitions can yield materially different pictures. We adopt JMP statistics as our benchmark because they are the reference used in official SDG reporting. Despite these validation constraints, the external-validation results indicate that the framework can recover meaningful large-scale patterns in infrastructure presence and capture national-scale variation across countries. The close agreement between model-derived and JMP piped-water estimates shows that fine-scale satellite predictions can be aggregated into population-weighted national estimates that align with official SDG 6 monitoring statistics. The sewage system results, while benchmarked against a less directly comparable sanitation indicator, still provide informative national-level agreement with a comparable degree of systematic bias. Together, these findings suggest that framework-derived estimates from satellite imagery and population data can complement conventional reporting systems, particularly in data-scarce settings, while also supporting the subnational analyses of infrastructure burden, deprivation severity, and environmental equity presented in the next section.


\subsection{Policy applications for SDG 6 monitoring and subnational infrastructure planning}
\subsubsection{Filling monitoring gaps in countries without survey coverage}
The policy value of this framework is particularly clear in countries where household survey data are entirely absent. Figure~\ref{fig:mae_survey_vs_nosurvey} compares population-weighted mean absolute error (MAE) between framework-derived estimates and JMP statistics for countries represented in Afrobarometer and countries without Afrobarometer survey coverage. The no-survey countries perform at least comparably to survey countries: for piped water, population-weighted MAE is lower in no-survey countries than in survey countries (9.5\% versus 12.9\%), while for sewage system access the errors are nearly identical (10.7\% versus 10.9\%).

Figure~\ref{fig:mae_survey_vs_nosurvey} further shows the share of the population in countries without Afrobarometer survey coverage whose estimates fall within different absolute-error thresholds. For piped water presence, no-survey countries collectively cover approximately 193.9 million people; approximately 62.6\% of this population (nearly 121.4 million people) resides in countries where model estimates fall within 15\% of JMP values, and nearly 60\% (115.7 million people) within 10\%. For safely managed sanitation, the no-survey countries represent 165.7 million people; estimates fall within 15\% of JMP values for 96.4\% of the population (159.7 million people). These results indicate that the framework can provide reasonably accurate population-level estimates for substantial populations living in countries without Afrobarometer survey coverage, highlighting its practical value for filling monitoring gaps where conventional survey-based evidence is unavailable. These figures, however, should be interpreted with caution. The no-survey evaluation covers only 11 countries for piped water and 6 for sewage (Table~\ref{tab:no_survey_country_breakdown}), and because the shares are weighted by population, they are dominated by one or two populous countries: a close match in a single large country can make the overall percentage appear high even when few countries contribute. These population-weighted percentages should therefore be read as indicative of a small set of countries, not as representative or generalizable estimates.

\subsubsection{Subnational burden and deprivation in infrastructure access in Nigeria}
Beyond extending monitoring coverage in countries without survey data, the framework can translate satellite-based predictions into practical evidence for SDG 6 planning. At 2.56 km spatial resolution, the model outputs reveal within-country variation in piped water and sewage systems presence that is obscured by national-level SDG statistics. These fine-scale estimates can serve as a screening tool to identify subnational regions with low predicted presence, helping governments and development partners prioritize investments, additional surveys, and field verification where they are most needed. By highlighting spatial disparities in system presence, the framework also supports more targeted infrastructure planning and equity monitoring. As an independent, top-down source of evidence derived from satellite imagery, it can complement bottom-up survey-based reporting systems and flag discrepancies that may warrant further validation and complement existing SDG tracking systems.

\begin{figure}[htbp]
  \centering
  \includegraphics[width=1\linewidth]{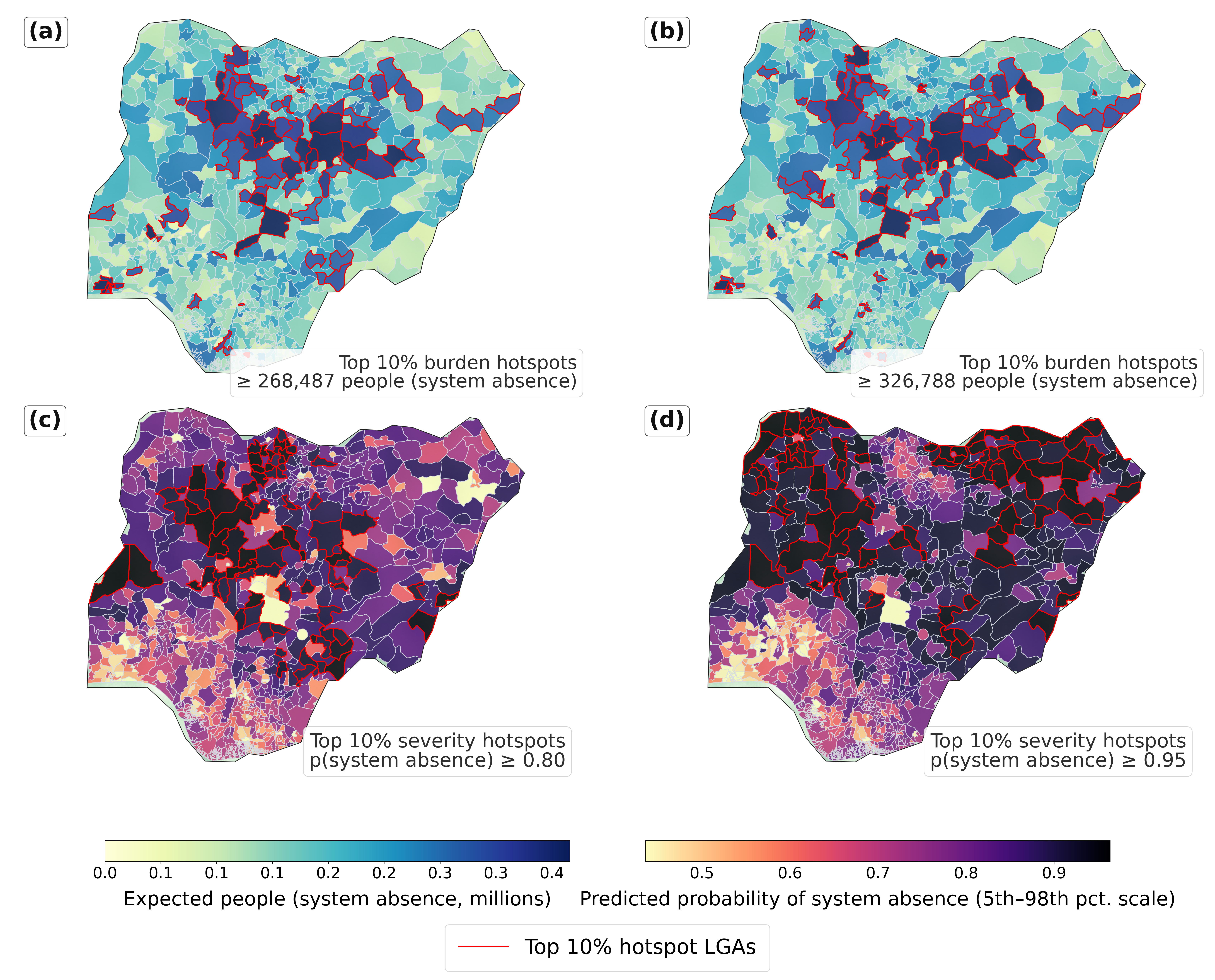}
  \caption{Nigeria LGA-level hotspot analysis. Panels (a) and (b) show population burden for piped water and sewage system presence, measured as the expected number of people living in areas with system absence. Panels (c) and (d) show deprivation severity, measured as the predicted probability of system absence. Red outlines mark hotspot LGAs, defined as the top 10\% of LGAs under the corresponding burden or severity metric. Separate color scales are used for the burden and severity panels.}
  \label{fig:nigeria_hotspots}
\end{figure}

To make these policy uses concrete, we applied the framework to Nigeria at the Local Government Area (LGA) level (Figure~\ref{fig:nigeria_hotspots}). Nigeria is a useful demonstration case because it is Africa's most populous country and has substantial subnational variation in urbanization, population density, and infrastructure presence. The analysis compares two complementary metrics: population burden, measured as the expected number of people living in areas without piped water or sewage system presence, and deprivation severity, measured as the predicted probability of system absence. In Figure~\ref{fig:nigeria_hotspots}, panels (a) and (b) show burden for piped water and sewage system presence, while panels (c) and (d) show severity for the same services; red outlines mark the top 10\% of LGAs under each metric.

The LGA-level results reveal substantial spatial inequality that would be obscured by national averages (Table~\ref{tab:nigeria_lga_metric_summary}). In the burden maps (Figures~\ref{fig:nigeria_hotspots}a and b), high-burden LGAs are concentrated across broad areas of central and northern Nigeria, with smaller concentrations in the southwest and parts of the south. These areas mark where large populations coincide with substantial predicted lack of system presence, making them especially relevant for interventions aimed at maximizing population reach. Quantitatively, the expected number of people living in areas with system absence averages 0.165 $\pm$ 0.090 million for piped water and 0.200 $\pm$ 0.107 million for sewage, but reaches 1.187 million and 1.577 million people in the highest-burden LGAs, respectively. These maxima are 8.1 and 8.7 times the median LGA burden, and the top 10\% of LGAs account for 21.1\% of total expected piped-water burden and 20.9\% of total expected sewage burden. This skew partly reflects the uneven population distribution across LGAs: mean LGA population is 0.265 $\pm$ 0.176 million, but the largest LGA contains 2.589 million people, or 11.4 times the median. Burden-based hotspots are therefore strongly shaped by population concentration and are most useful for identifying where investments could reach the largest number of people.

\begin{table}[htbp]

\centering

\caption{LGA-level variation in predicted piped water and sewage system presence in Nigeria. Burden is reported as expected people living in areas with system absence in millions, and severity as the predicted probability of system absence. The max/median ratio compares the most affected LGA with a typical LGA, while the top 10\% share shows how much of the total burden is concentrated in the top 10\% of LGAs. Values are reported as mean $\pm$ one standard deviation across the 767 LGAs. Values use isotonic-calibrated patch-level probabilities.}

\label{tab:nigeria_lga_metric_summary}

\resizebox{\textwidth}{!}{%
\begin{tabular}{llrrrrrr}
\toprule

\textbf{Metric} & \textbf{Unit} & \textbf{Mean $\pm$ SD} & \textbf{Median} & \textbf{90th pct.} & \textbf{Max} & \textbf{Max/median} & \textbf{Top 10\% share} \\

\midrule
Piped water burden & million people & 0.165 $\pm$ 0.090 & 0.147 & 0.268 & 1.187 & 8.1 & 21.1\% \\
Sewage system burden & million people & 0.200 $\pm$ 0.107 & 0.181 & 0.327 & 1.577 & 8.7 & 20.9\% \\
Piped water severity & probability & 0.645 $\pm$ 0.129 & 0.667 & 0.795 & 0.904 & 1.4 & 12.9\% \\
Sewage system severity & probability & 0.785 $\pm$ 0.155 & 0.801 & 0.952 & 0.983 & 1.2 & 12.3\% \\
LGA population & million people & 0.265 $\pm$ 0.176 & 0.227 & 0.422 & 2.589 & 11.4 & 23.4\% \\
\bottomrule

\end{tabular}%

}

\end{table}

Severity, by contrast, shows a different pattern of inequality, as illustrated in Figures~\ref{fig:nigeria_hotspots}c and d. Predicted absence probability averages 0.645 $\pm$ 0.129 for piped water and 0.785 $\pm$ 0.155 for sewage systems, indicating that deprivation is widespread rather than confined to a small number of extreme LGAs. This broader pattern is visible in the severity maps, where high absence probabilities extend across many LGAs, particularly in northern Nigeria. The distributional metrics support this interpretation: maximum-to-median ratios are relatively constrained at 1.4 for piped water and 1.2 for sewage, and the top-decile shares are 12.9\% and 12.3\%, respectively. This pattern is especially pronounced for sewage systems, where the 90th-percentile absence probability is 0.952, indicating that LGAs classified as severity hotspots are highly likely to have system absence.

Because the LGA-level burden estimates were obtained by aggregating calibrated patch-level probabilities over population, we assessed probability calibration and quantified uncertainty in the resulting burden estimates. Held-out reliability analysis showed that the raw $k$-NN probabilities were already reasonably well calibrated, with expected calibration errors of 0.023 for piped water and 0.018 for sewage systems (Table~\ref{tab:calibration}). We nevertheless applied isotonic recalibration, which approximately halved the calibration error for piped water. We then used bootstrap resampling to characterize variability in the LGA-level burden estimates under repeated tile resampling and calibration-model refitting. In each bootstrap replicate, tiles were resampled with replacement within each LGA, the isotonic calibration model was refitted, and the population-weighted burdens and associated summary statistics were recalculated. Based on the calibrated probabilities and bootstrap distributions, the highest LGA-level burdens were 1.187 million people living in areas without piped water (95\% bootstrap CI: 0.87--1.48 million) and 1.577 million people living in areas without sewage systems (95\% bootstrap CI: 1.36--1.80 million). The corresponding median LGA-level burdens were 0.147 million (95\% bootstrap CI: 0.143--0.151 million) and 0.180 million (95\% bootstrap CI: 0.175--0.185 million), respectively, while the ratios of the maximum to the mean burden were 7.2 (95\% bootstrap CI: 5.7--9.0) and 7.9 (95\% bootstrap CI: 6.8--9.0). Together, these results indicate that the estimated burden patterns were generally stable across bootstrap replicates. The top-decile burden shares, at 21.1\% for piped water and 20.9\% for sewage systems, and the corresponding severity thresholds, at 0.795 and 0.952, respectively, showed negligible variation across bootstrap replicates, further supporting the robustness of the identified high-burden and high-severity areas.

These results show why burden and severity should be interpreted together. Burden identifies where investments could reach the largest number of affected people, while severity identifies where system absence is most pervasive, including less populous LGAs that could be missed by population-weighted prioritization alone. From a planning perspective, LGAs that rank highly on both metrics may warrant particular attention because they combine large affected populations with systematic service deprivation. Used together, the two metrics provide a more balanced basis for SDG 6 planning: maximizing population reach while also prioritizing areas where system absence is most pervasive.

\subsection{Limitations}

The most important limitation concerns the unit of observation. Our labels are enumerator observations of whether a piped water system or sewage system is present in the surveyed enumeration area, not household reports of service use. In Afrobarometer Round 8, agreement between the area-level piped water label and household-reported piped water use is 74.3\%: the label identifies absence reliably (90.7\% of households where no system is recorded report no piped supply) but overstates household use (61.0\% of households where a system is recorded report piped water into dwelling or yard). Our estimates should therefore be read as maps of where water and sanitation systems exist, which bound but do not measure the household-use quantities defined by SDG Indicators 6.1.1 and 6.2.1. Satellite-derived estimates are a complement to household survey measurement.

Another limitation is that the availability of high-quality satellite
imagery, such as cloud-free imagery at adequate spatial resolution,
varies across regions and may introduce geographic bias if some areas
systematically have better image coverage than others. Moreover,
Afrobarometer household survey data, while extensive, do not
represent every local context and may therefore introduce uncertainty when
extrapolating to unsampled countries or to the continental scale.
In addition, decentralized water supply and sanitation systems
may be difficult to infer from optical satellite imagery,
particularly where visible built-environment proxies are weak.
\section{Conclusions}

Meeting SDG Targets 6.1 and 6.2 by 2030 will require not only a six- to eightfold acceleration in coverage expansion~\citep{WHOJMP2025}, but also a shift toward more timely and spatially detailed systems for monitoring infrastructure. This study presents one such complementary approach: we show that fine-scale signals of piped water and sewage system presence can be inferred from freely available Sentinel-2 imagery using DINO-based self-supervised visual representations and a lightweight $k$-nearest-neighbor classifier.

Across held-out Afrobarometer survey locations, the best-performing model achieves AUROC values of 91.54\% for piped water and 93.24\% for sewage system presence. When applied across 50 African countries at approximately 2.56 km spatial resolution, the resulting population-weighted national estimates closely track JMP statistics for piped water access ($R^2 = 0.92$). Agreement is weaker for sewage system relative to JMP sanitation statistics ($R^2 = 0.72$), but this comparison is constrained by a smaller validation sample ($n=27$ versus $n=48$ for piped water) and by the nearest available but broader JMP sanitation benchmark, which includes facility types beyond sewage systems. The similar fitted slopes for sewage system and piped water nevertheless suggest comparable systematic bias despite this benchmark mismatch.

Most notably, in countries without Afrobarometer survey coverage, the framework achieves population-weighted MAEs of 9.5\% for piped water and 10.7\% for sewage across populations of 193.9 million and 165.7 million people, respectively; estimates fall within 15\% of JMP values for 121.4 million people for piped water and 159.7 million people for sanitation. These results indicate that satellite-derived estimates can provide robust out-of-distribution performance in precisely the geographies where conventional survey coverage is most limited.

The Nigeria application further illustrates how spatially disaggregated predictions from our framework can make SDG 6 monitoring more actionable by identifying subnational areas where infrastructure absence and population burden are greatest. Across 767 LGAs, the expected number of people without infrastructure reaches 1.187 million for piped water and 1.577 million for sewage in the highest-burden LGAs, corresponding to 8.1 and 8.7 times the median LGA burden, respectively. The top 10\% of LGAs account for 21.1\% of total expected piped-water burden and 20.9\% of total expected sewage burden, highlighting areas where additional surveys, field verification, and infrastructure planning could be prioritized. At the same time, top-decile absence probability thresholds of 0.795 for piped water and 0.952 for sewage show that predicted system absence is widespread across many LGAs, not confined to a few extreme cases. Used together, these two metrics help identify LGAs that may warrant priority follow-up and planning: places where large affected populations coincide with pervasive service predicted system absence.

The economics of the approach also warrant emphasis. Sentinel-2 imagery is freely and continuously available, reducing the marginal cost of refreshing national estimates to near zero relative to the \$300-per-household surveys currently sustaining SDG~6 tracking. Taken together, this work suggests that Earth observation can provide a scalable, low-latency layer of spatially detailed evidence on where water and sanitation systems exist, refreshable continuously between survey rounds and useful for targeting household surveys, field verification, infrastructure planning, and equity assessment. By complementing household surveys, which remain essential for measuring the service-use dimensions defined by SDG Indicators 6.1.1 and 6.2.1, this framework can help make SDG 6 monitoring more timely, spatially detailed, and actionable. More broadly, the framework could be transferred to other regions with comparable multispectral satellite coverage and baseline ground-reference data, and adapted to other spatially observable sustainable development indicators, including energy infrastructure, urban expansion, and land productivity. This study underscores the broader value of integrating satellite-derived inference into WASH and SDG data systems as a practical tool for identifying communities where infrastructure gaps and population burdens are greatest, prioritizing survey follow-up, and supporting local-scale infrastructure planning toward more equitable sustainable development.


\section*{Code and Data Availability}

The code used for data preprocessing, self-supervised pretraining, held-out evaluation, and external validation is publicly available on GitHub at \href{https://github.com/othmaneechc/SDG6-Tracker}{https://github.com/othmaneechc/SDG6-Tracker}.

The data and model artifacts associated with this study are archived on Zenodo. Pretrained DINO and DINOv2 weights, inference results, and population density patches are available at \href{https://zenodo.org/records/19156085}{https://zenodo.org/records/19156085}.

Afrobarometer imagery tiles are available at \href{https://zenodo.org/records/14740420}{https://zenodo.org/records/14740420}.

\section*{Acknowledgment}

The authors acknowledge funding support from the DKU Summer Research Program Grant.

\bibliographystyle{apalike}
\bibliography{references}

\clearpage

\appendix

\clearpage
\setcounter{page}{1}
\pagenumbering{arabic}

\begin{center}

{\Large\bfseries
Supplementary Material

Seeing SDG 6 from space: local-scale monitoring of piped water and sewage systems across Africa using satellite imagery and self-supervised learning\par}

\vspace{3em}

{\normalsize
Othmane Echchabi$^{1,2,5, \dagger}$, Aya Lahlou$^{3,4,5,\dagger}$, Nizar Talty$^{5}$, Josh Malcolm Manto$^{5}$, Tongshu Zheng$^{5,*}$, and Ka Leung Lam$^{5,*}$\par}
\end{center}

\vspace{0.8em}

{\small
$^{1}$ Mila -- Quebec AI Institute\par
$^{2}$ School of Computer Science, McGill University\par
$^{3}$ Department of Earth and Environmental Engineering, Columbia University\par
$^{4}$ Center for Learning the Earth with Artificial Intelligence and Physics (LEAP)\par
$^{5}$ Division of Natural and Applied Sciences, Duke Kunshan University\par}

\vspace{0.5em}

{\footnotesize
$^{*}$ Corresponding authors: \href{mailto:tongshu.zheng@dukekunshan.edu.cn}{tongshu.zheng@dukekunshan.edu.cn}; \href{mailto:kaleung.lam@dukekunshan.edu.cn}{kaleung.lam@dukekunshan.edu.cn}\par
$^{\dagger}$ These authors contributed equally to this work.\par}

\vspace{1em}


\renewcommand{\thesection}{\Roman{section}}
\renewcommand{\thesubsection}{\thesection.\arabic{subsection}}

\renewcommand{\thefigure}{S\arabic{figure}}
\renewcommand{\thetable}{S\arabic{table}}

\setcounter{section}{0}
\setcounter{subsection}{0}
\setcounter{figure}{0}
\setcounter{table}{0}

\section{Visual Examples}
\label{sec:appendix_visual_examples}

\begin{figure}[!htbp]
    \centering

    {\textbf{Regions without piped water system presence}}\\[0.8em]

    \begin{subfigure}[t]{0.3\textwidth}
        \centering
        \includegraphics[width=\linewidth,height=4cm]{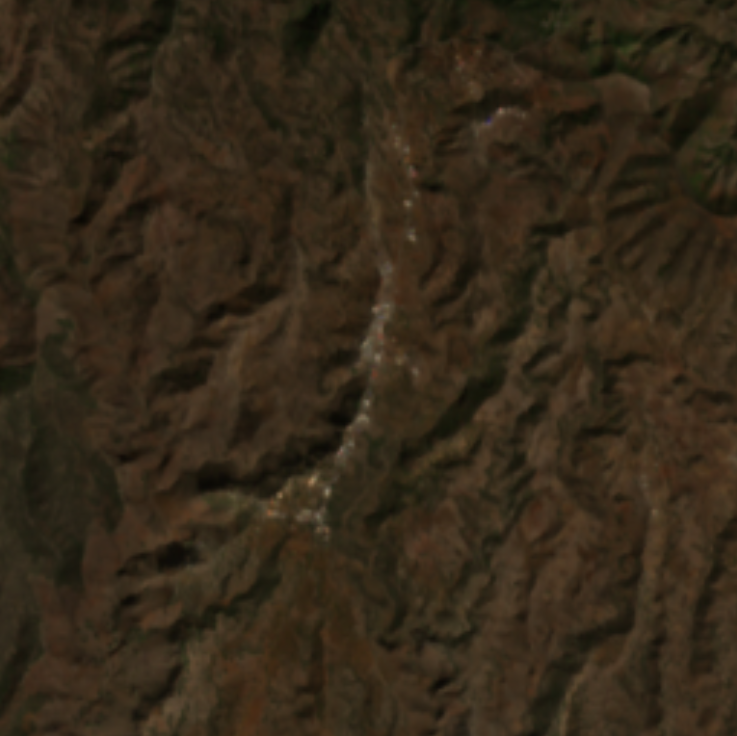}
        \caption*{15$^{\circ}$01'03.4"N 23$^{\circ}$36'58.1"W\\Cabo Verde}
    \end{subfigure}
    \hfill
    \begin{subfigure}[t]{0.3\textwidth}
        \centering
        \includegraphics[width=\linewidth,height=4cm]{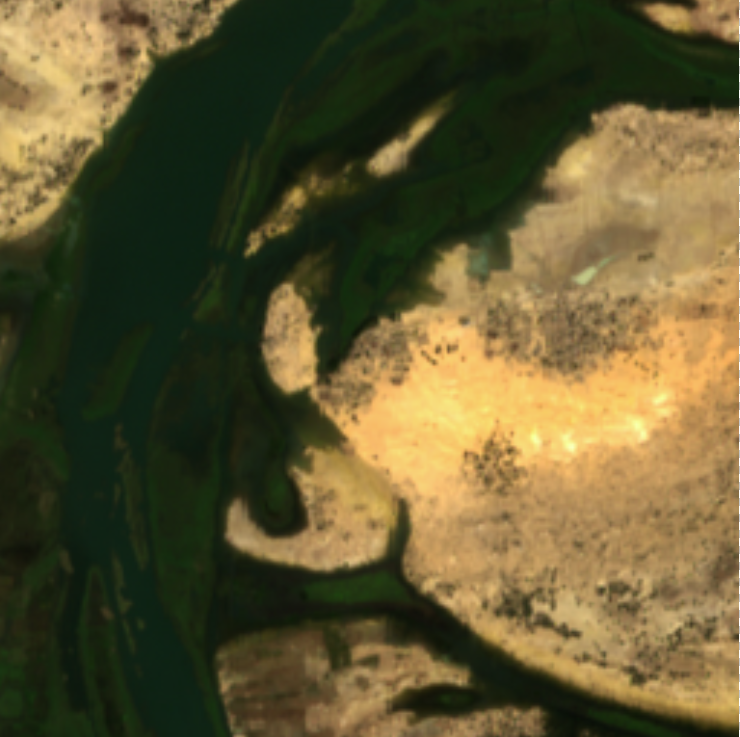}
        \caption*{16$^{\circ}$13'33.6"N 3$^{\circ}$13'28.2"W\\Mali}
    \end{subfigure}
    \hfill
    \begin{subfigure}[t]{0.3\textwidth}
        \centering
        \includegraphics[width=\linewidth,height=4cm]{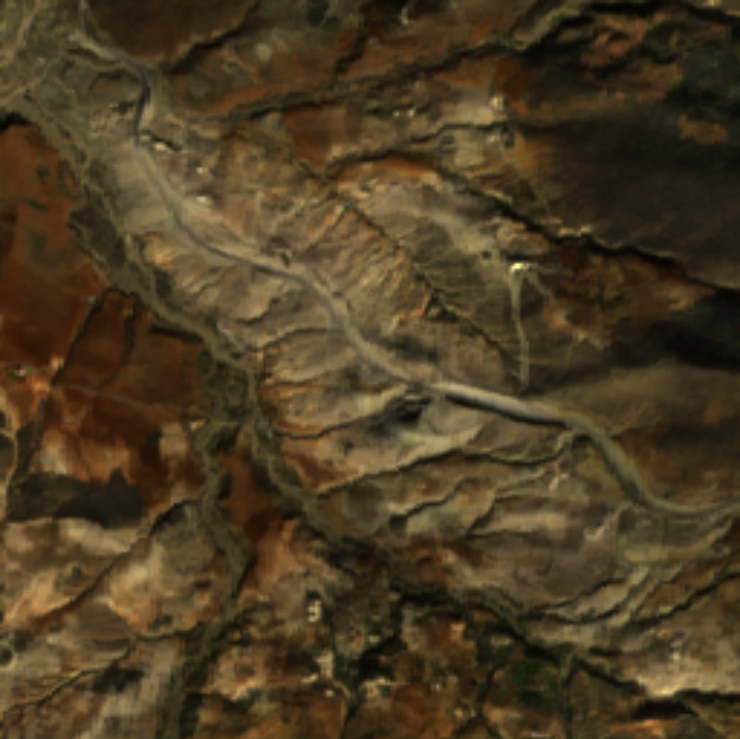}
        \caption*{35$^{\circ}$07'52.7"N 4$^{\circ}$30'55.5"W\\Morocco}
    \end{subfigure}

    \vspace{1.2em}

    {\textbf{Regions with piped water system presence}}\\[0.8em]

    \begin{subfigure}[t]{0.3\textwidth}
        \centering
        \includegraphics[width=\linewidth,height=4cm]{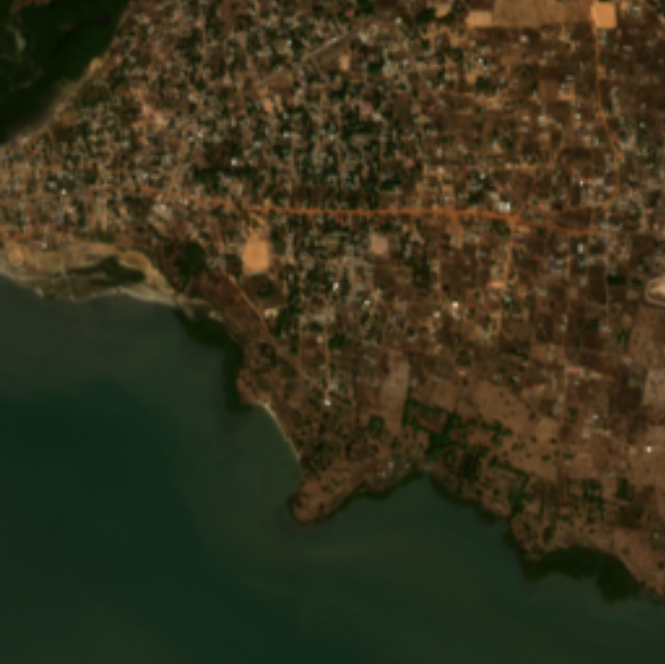}
        \caption*{13$^{\circ}$28'41.9"N 16$^{\circ}$31'37.3"W\\The Gambia}
    \end{subfigure}
    \hfill
    \begin{subfigure}[t]{0.3\textwidth}
        \centering
        \includegraphics[width=\linewidth,height=4cm]{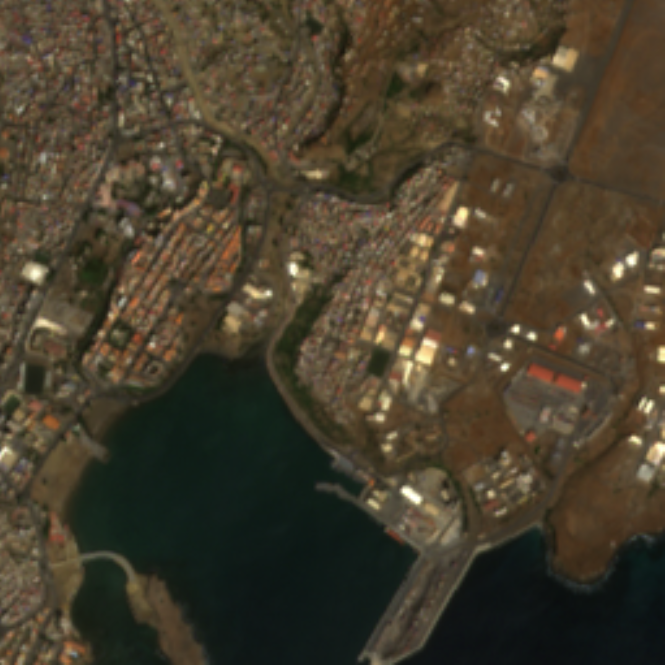}
        \caption*{14$^{\circ}$55'03.8"N 23$^{\circ}$30'06.3"W\\Cabo Verde}
    \end{subfigure}
    \hfill
    \begin{subfigure}[t]{0.3\textwidth}
        \centering
        \includegraphics[width=\linewidth,height=4cm]{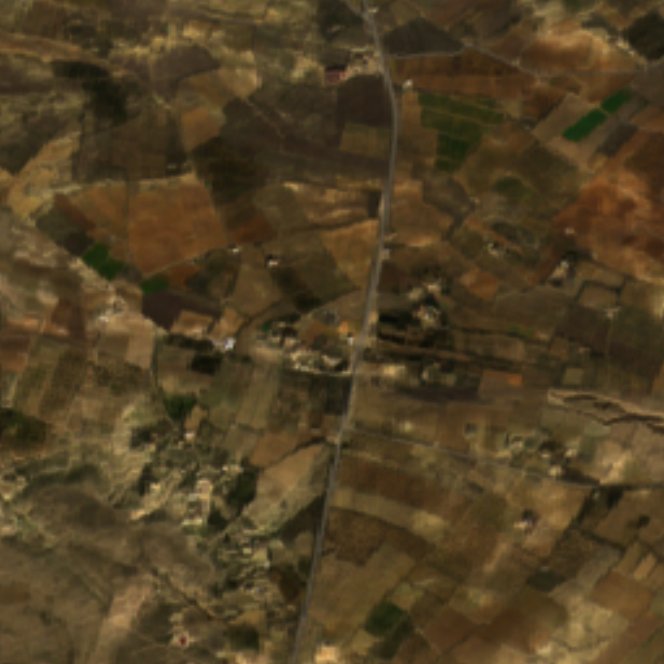}
        \caption*{35$^{\circ}$24'59.6"N 9$^{\circ}$07'27.9"E\\Tunisia}
    \end{subfigure}

    \caption{Examples of Sentinel-2 patches from locations without and with reported piped water system presence. These examples illustrate how infrastructure presence may be associated with visible built-environment and land-use patterns even when the infrastructure itself is not directly observable.}
    \label{fig:piped_water_examples}
\end{figure}

Because piped water and sewage infrastructure are not always directly visible in Sentinel-2 imagery, model predictions likely rely on indirect built-environment signals such as settlement density, road structure, land-use patterns, and urban form. Figure~\ref{fig:piped_water_examples} provides examples of image patches from locations with and without reported piped water presence.

\FloatBarrier

\section{Methodological Details}
\label{sec:appendix_methods}

\begin{figure}[!htbp]
    \centering
    \includegraphics[width=0.7\linewidth]{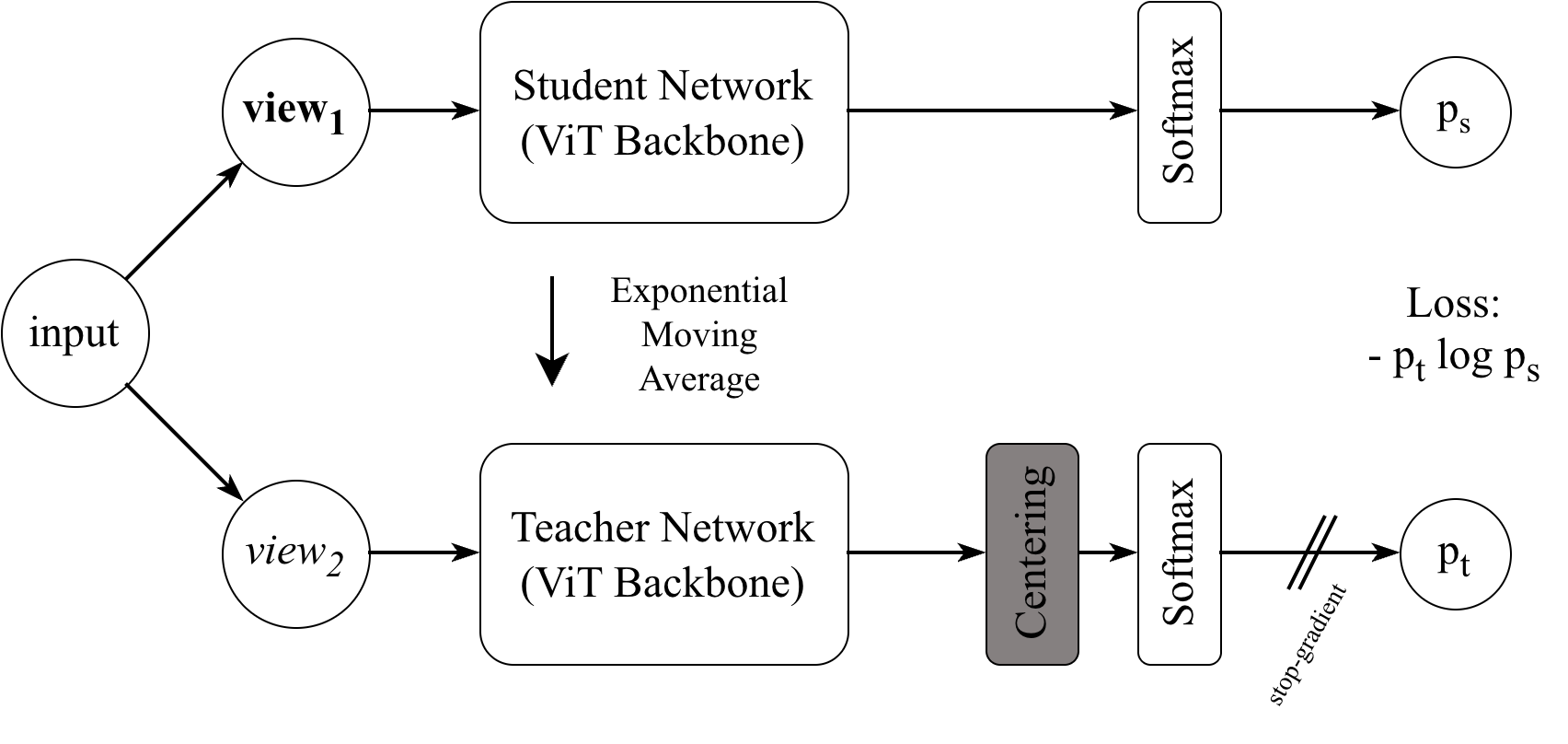}
    \caption{Overview of the DINO self-distillation framework \citep{Caron2021}. Two augmented views of the same image are processed by a student and a teacher ViT backbone. The student prediction $p_s$ is optimized to match the teacher prediction $p_t$ via a cross-entropy loss, while the teacher is updated as an exponential moving average of the student.}
    \label{fig:dino_framework}
\end{figure}

This section provides additional details on the self-supervised representation-learning pipeline and held-out evaluation setup. Figure~\ref{fig:dino_framework} summarizes the DINO self-distillation framework, and Table~\ref{tab:training_details} reports the main pretraining and evaluation settings.

\begin{table}[!htbp]
\centering
\caption{Implementation details for pretraining and held-out evaluation.}
\label{tab:training_details}
\small
\begin{tabular}{p{2.2cm}p{2.2cm}p{9.4cm}}
\toprule
Stage & Model & Key settings \\
\midrule
Pretraining & DINO &
8 NVIDIA A40 GPUs; distributed training; ViT-Base; patch size $8$; 300 epochs; batch size $16$ per GPU; learning rate $3\times 10^{-4}$; mixed precision (FP16). \\

Pretraining & DINOv2 &
8 NVIDIA A40 GPUs; distributed training; ViT-Large; patch size $8$; 200 epochs; 20 warmup epochs; batch size $16$ per GPU; centering set to Sinkhorn-Knopp; training configuration defined by project overrides together with official DINOv2 default settings. \\

Held-out evaluation & DINO &
Teacher checkpoint used for embedding extraction; $k \in \{5,10,20,50,100,200\}$; cosine similarity; softmax-weighted $k$-NN; temperature $0.07$; batch size $64$; 4 workers. \\

Held-out evaluation & DINOv2 &
Teacher checkpoint used for embedding extraction; $k \in \{5,10,20,50,100,200\}$ during model selection; cosine similarity; softmax-weighted $k$-NN; temperature $0.07$; batch size $128$; 4 workers; resize $256$ and crop $224$. \\

Held-out evaluation & DINOv3 &
Public pretrained checkpoint; $k \in \{5,10,20,50,100,200\}$; cosine similarity; softmax-weighted $k$-NN; temperature $0.07$; batch size $64$; 4 workers. \\

Held-out evaluation & Galileo &
Public pretrained checkpoint; $k \in \{5,10,20,50,100,200\}$; cosine similarity; softmax-weighted $k$-NN; temperature $0.07$; batch size $64$; 4 workers; bf16 inference; bands B2/B3/B4. \\

Held-out evaluation & Prithvi &
Public pretrained checkpoint; $k \in \{5,10,20,50,100,200\}$; cosine similarity; softmax-weighted $k$-NN; temperature $0.07$; batch size $32$; 4 workers; resize+center crop $224$; RGB channel indices $[0,1,2]$. \\
\bottomrule
\end{tabular}
\end{table}

\FloatBarrier
\section{Additional Foundation Model Results}
\label{sec:appendix_foundation_models}

Table~\ref{tab:foundation_model_summary} reports supplementary held-out results for additional Earth observation foundation models. These results complement the main model-comparison table and show that pretrained geospatial encoders differ substantially in their transferability to infrastructure-access prediction.

\begin{table}[!htbp]
\centering
\caption{Additional held-out test-set results for selected foundation models on piped water and sewage system access. For each model--task pair, the value of $k$ was selected based on validation performance, and the corresponding held-out test metrics are reported in \%.}

\label{tab:foundation_model_summary}
\small
\resizebox{\textwidth}{!}{%
\setlength{\tabcolsep}{5pt}
\begin{tabular}{llccccccc}
\toprule
\textbf{Task} & \textbf{Model} & \textbf{Backbone} & \textbf{Pretraining data} & \textbf{Params (M)} & \textbf{$k$} & \textbf{Accuracy} & \textbf{Recall} & \textbf{F1} \\
\midrule
\multirow{5}{*}{Piped water}
& DINOv3 & ViT-B & LVD-1689M & 86 & 10 & 80.65 & 80.69 & 80.49 \\
& DINOv3 & ViT-L & SAT-493M & 300 & 10 & \textbf{81.43} & \textbf{81.22} & \textbf{81.19} \\
& Galileo & ViT-B & Galileo-pretrain & 86 & 10 & 76.97 & 76.44 & 76.54 \\
& Prithvi-EO-2.0 & ViT-L & HLS time series & 300 & 10 & 77.91 & 77.71 & 77.64 \\
& Prithvi-EO-2.0 & ViT-H & HLS time series & 600 & 10 & 77.69 & 77.30 & 77.34 \\
\midrule
\multirow{5}{*}{Sewage system}
& DINOv3 & ViT-B & LVD-1689M & 86 & 20 & \textbf{84.85} & \textbf{81.32} & \textbf{82.16} \\
& DINOv3 & ViT-L & SAT-493M & 300 & 10 & 84.65 & 81.06 & 81.92 \\
& Galileo & ViT-B & Galileo-pretrain & 86 & 10 & 81.69 & 77.77 & 78.48 \\
& Prithvi-EO-2.0 & ViT-L & HLS time series & 300 & 10 & 82.88 & 79.05 & 79.84 \\
& Prithvi-EO-2.0 & ViT-H & HLS time series & 600 & 10 & 82.59 & 79.22 & 79.73 \\
\bottomrule
\end{tabular}%
}
\end{table}

Galileo and Prithvi-EO-2.0 consistantly scored lower than DINO, DINOv2, and DINOv3 in our setup. These models were, however, evaluated on the same uint8 RGB visualization patches used for the DINO family, which lie outside their expected input domain: both were pretrained on multispectral surface reflectance. Their scores should therefore be interpreted as reflecting a preprocessing and input-domain mismatch rather than an intrinsic limitation of the models; a fair comparison would require re-exported raw multispectral imagery.

\FloatBarrier

\section{Robustness Analyses: Spatial Validation, Calibration, and Confounding}
\label{sec:appendix_robustness}

\begin{figure}[!htbp]
  \centering
  \includegraphics[width=0.78\linewidth]{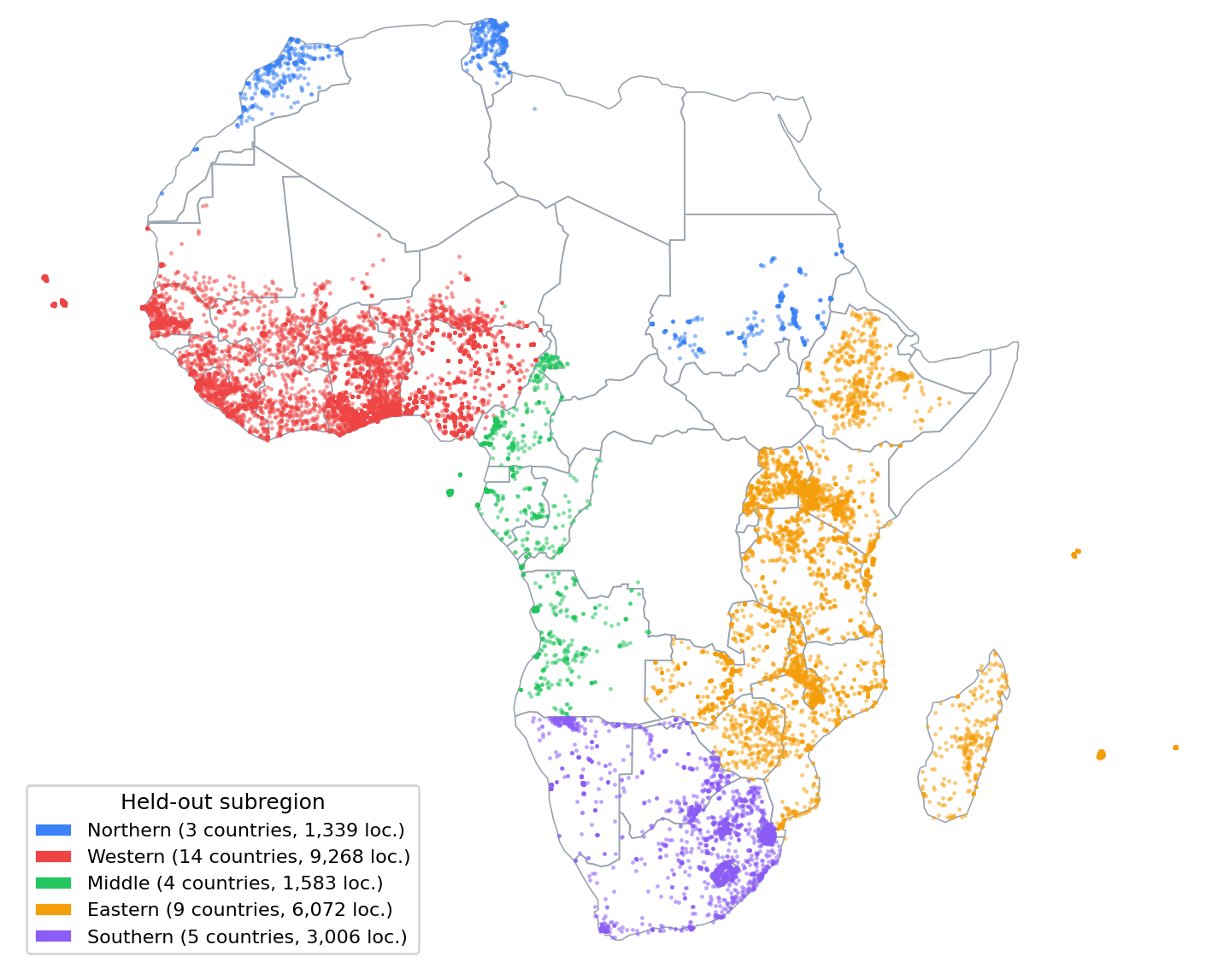}
  \caption{Leave-one-region-out cross-validation folds: survey locations colored by the UN African subregion used as the held-out fold. Stricter than a random split because entire regions are absent from training.}
  \label{fig:region_folds_cv}
\end{figure}

This section reports additional robustness analyses: generalization under alternative data splits (Table~\ref{tab:split_schemes}), the leave-one-region-out fold construction (Figure~\ref{fig:region_folds_cv}), probability calibration (Table~\ref{tab:calibration}), and settlement-stratified evaluation of the urban--rural and population-density confounds (Table~\ref{tab:stratified}). All use the DINOv2 embeddings at $k=200$; cross-validation folds are formed over survey locations or over whole subregions, never over individual images.

The leave-one-region-out folds assign each survey country to one of the five UN African subregions and hold out one subregion at a time. Because entire macro-regions -- and all spatially autocorrelated neighbours within them -- are removed from training, this is a substantially stricter test of geographic transfer than a random location split.

\begin{table}[!htbp]
\centering
\caption{Held-out AUROC (\%) under alternative evaluation splits, DINOv2 at $k=200$; mean~$\pm$~s.d.\ across folds. Random 5-fold shuffles survey locations; leave-one-region-out holds out each of the five UN African subregions (Northern, Western, Middle, Eastern, Southern) in turn, so an entire macro-region is absent from training (fold assignment in Figure~\ref{fig:region_folds_cv}). Random splitting matches the original split, but holding out a whole subregion lowers AUROC by roughly 14--16 points, quantifying the difficulty of transfer to geographically disjoint regions.}

\label{tab:split_schemes}
\small
\begin{tabular}{lcc}
\toprule
\textbf{Evaluation split} & \textbf{Piped water} & \textbf{Sewage} \\
\midrule
Original (80/10/10) & 91.54 & 93.24 \\
Random 5-fold & 91.26 $\pm$ 0.39 & 92.47 $\pm$ 0.40 \\
Leave-one-region-out (5 UN subregions) & 75.46 $\pm$ 3.64 & 78.67 $\pm$ 5.58 \\
\bottomrule
\end{tabular}
\end{table}

\begin{table}[!htbp]
\centering
\caption{Probability calibration on the held-out test set, DINOv2 at $k=200$. Calibrators are fit on validation and evaluated on test. Temperature scaling applies one logit-scale parameter, Platt scaling fits an affine logit transform, isotonic regression fits a monotone non-parametric mapping, and the logistic head is trained directly on embeddings. The ranking-preserving methods leave AUROC essentially unchanged; isotonic regression roughly halves the piped-water expected calibration error (ECE) at negligible AUROC cost and is adopted, whereas the logistic head is worse on both axes. Brier and ECE: lower is better.}
\label{tab:calibration}
\small
\begin{tabular}{lcccccc}
\toprule
& \multicolumn{3}{c}{\textbf{Piped water}} & \multicolumn{3}{c}{\textbf{Sewage}} \\
\cmidrule(lr){2-4}\cmidrule(lr){5-7}
\textbf{Method} & AUROC & Brier & ECE & AUROC & Brier & ECE \\
\midrule
None ($k$-NN) & 91.54 & 0.115 & 0.023 & 93.24 & 0.094 & 0.018 \\
Temperature & 91.54 & 0.116 & 0.029 & 93.24 & 0.094 & 0.019 \\
Platt & 91.54 & 0.116 & 0.028 & 93.24 & 0.094 & 0.017 \\
Isotonic & 91.48 & 0.115 & 0.013 & 93.21 & 0.094 & 0.016 \\
Logistic head & 87.01 & 0.149 & 0.063 & 89.01 & 0.126 & 0.059 \\
\bottomrule
\end{tabular}
\end{table}

\begin{table}[!htbp]
\centering
\caption{Confound controls: held-out AUROC (\%) under random 5-fold cross-validation over survey locations (DINOv2 at $k{=}200$; mean $\pm$ s.d.\ across folds). Each baseline is fit on the training folds and applied to the held-out fold; population density is the nearest Meta grid tile population (available for 94\% of locations; Morocco and Sudan lack a grid file). The image embeddings exceed the strongest confounder baseline by 13.3--14.3 points and retain 85.9--88.2\% AUROC \emph{within} urban and rural strata, where the urban/rural indicator is constant (0.5) by construction. We drop samples with invalid \texttt{URBRUR} codes.}
\label{tab:stratified}
\small
\begin{tabular}{lcc}
\toprule
\textbf{Predictor} & \textbf{Piped water} & \textbf{Sewage system} \\
\midrule
Urban/rural indicator only & 75.07 $\pm$ 0.73 & 77.66 $\pm$ 0.73 \\
Population density only & 68.76 $\pm$ 1.07 & 70.40 $\pm$ 1.19 \\
Urban/rural {+} population density & 76.85 $\pm$ 1.04 & 79.05 $\pm$ 1.11 \\
Image embeddings (DINOv2) & \textbf{91.17 $\pm$ 0.19} & \textbf{92.34 $\pm$ 0.72} \\
\midrule
\multicolumn{3}{l}{\emph{Image embeddings within settlement strata:}} \\
\quad Urban locations only & 88.13 $\pm$ 0.68 & 88.15 $\pm$ 1.27 \\
\quad Rural locations only & 85.91 $\pm$ 0.48 & 87.90 $\pm$ 1.44 \\
\bottomrule
\end{tabular}
\end{table}

\FloatBarrier

\section{Additional External-Validation Results}
\label{sec:appendix_external}

\begin{figure}[!htbp]
  \centering
  \includegraphics[width=0.6\linewidth]{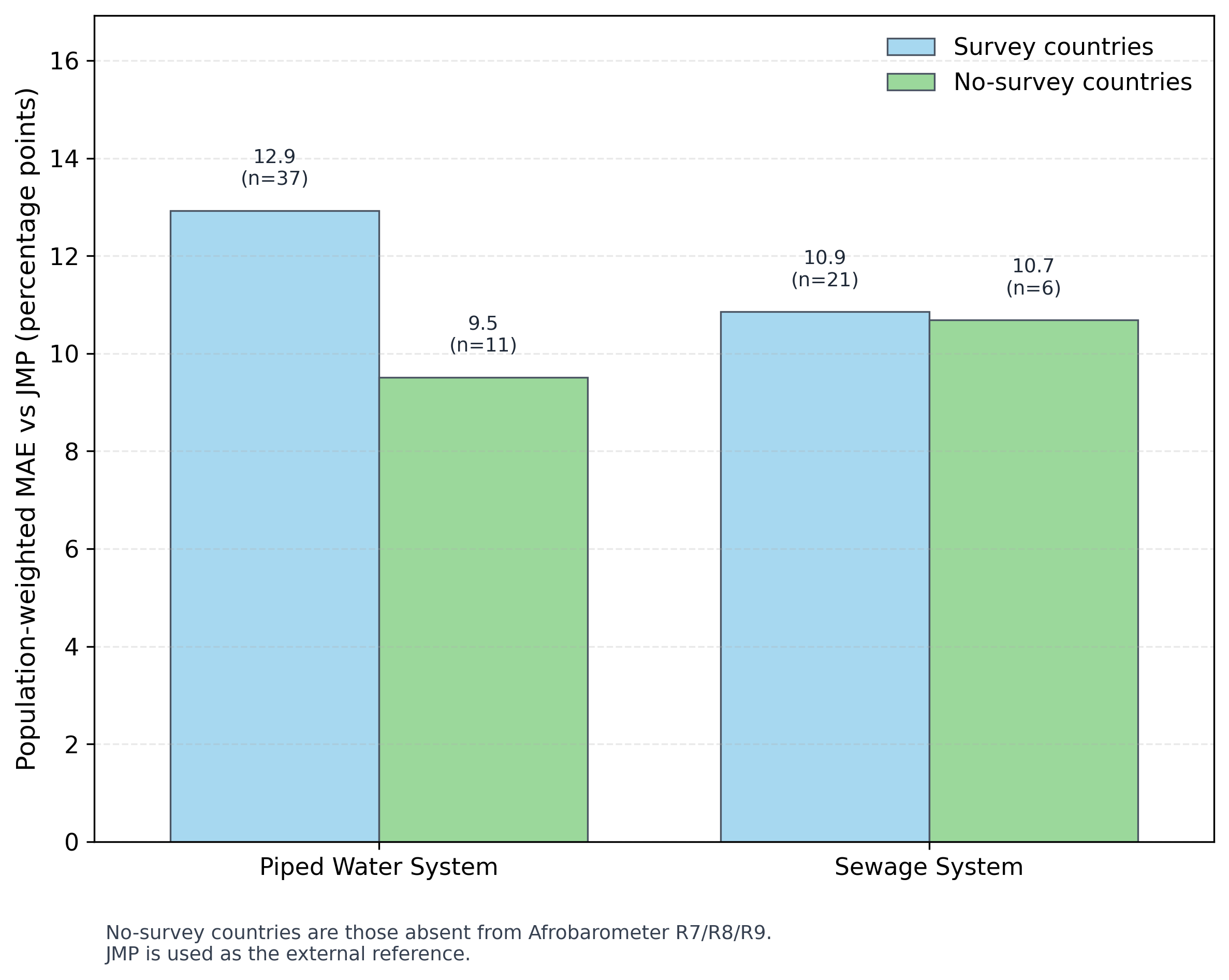}
  \caption{Population-weighted mean absolute error (MAE, \%) between model-derived estimates and JMP statistics for countries represented in Afrobarometer survey data and countries without Afrobarometer survey coverage. Results are shown for piped water and sewage system access. Numbers above bars indicate MAE values, and $n$ denotes the number of countries in each group.}
  \label{fig:mae_survey_vs_nosurvey}
\end{figure}

This section provides supplementary external-validation results for countries without Afrobarometer survey coverage. Figure~\ref{fig:mae_survey_vs_nosurvey} compares population-weighted MAE between survey-covered countries and countries without Afrobarometer survey coverage, while Table~\ref{tab:no_survey_country_breakdown} reports country-level results.

The results for countries without Afrobarometer survey coverage show that the model can provide useful estimates beyond countries represented in the training and held-out evaluation data. However, errors vary by country, likely reflecting differences in geographic transferability, benchmark definitions, and how strongly visible settlement patterns correspond to infrastructure access.

\begin{table}[!htbp]
\centering
\caption{Country-level external-validation results for countries without Afrobarometer survey coverage. Absolute error is computed as the absolute difference between the model-derived estimate and the corresponding JMP national statistic. Here, pp denotes percentage points.}
\label{tab:no_survey_country_breakdown}
\small
\resizebox{\textwidth}{!}{%
\begin{tabular}{llrrrr}
\toprule
\textbf{Task} & \textbf{Country} & \textbf{Population (M)} & \textbf{Prediction (\%)} & \textbf{JMP (\%)} & \textbf{Abs. error (pp)} \\
\midrule
Piped water & Egypt & 98.9 & 97.82 & 98.69 & 0.87 \\
Piped water & Chad & 16.8 & 25.17 & 16.79 & 8.38 \\
Piped water & Libya & 5.7 & 92.46 & 77.96 & 14.50 \\
Piped water & Burundi & 11.4 & 20.13 & 38.67 & 18.54 \\
Piped water & Algeria & 42.0 & 90.54 & 71.39 & 19.15 \\
Piped water & Rwanda & 12.1 & 20.31 & 45.49 & 25.18 \\
Piped water & Guinea-Bissau & 2.0 & 21.18 & 47.81 & 26.63 \\
Piped water & Eritrea & 3.4 & 21.73 & 51.76 & 30.03 \\
Piped water & Comoros & 0.9 & 25.06 & 59.50 & 34.44 \\
Piped water & Djibouti & 0.3 & 38.42 & 83.36 & 44.94 \\
Piped water & Mayotte & 0.3 & 43.78 & 92.67 & 48.89 \\
\midrule
Sewage system & Chad & 16.8 & 2.43 & 10.91 & 8.48 \\
Sewage system & Algeria & 42.0 & 52.98 & 62.41 & 9.43 \\
Sewage system & Egypt & 98.9 & 57.02 & 67.17 & 10.15 \\
Sewage system & Guinea-Bissau & 2.0 & 0.56 & 15.44 & 14.88 \\
Sewage system & Libya & 5.7 & 56.78 & 23.83 & 32.95 \\
Sewage system & Djibouti & 0.3 & 4.11 & 39.60 & 35.49 \\
\bottomrule
\end{tabular}%
}
\end{table}

\FloatBarrier

\section{Survey Context}
\label{sec:appendix_survey_context}

Table~\ref{tab:urban_rural_summary} reports mean area-level system-presence rates by survey round and urbanicity at the enumeration-area level. Urban enumeration areas have consistently higher presence to both piped water and sewage systems than rural areas, supporting the limitation that model predictions may partly reflect urban--rural settlement patterns rather than direct observation of infrastructure.

\begin{table}[!htbp]
\centering
\caption{Mean area-level system-presence rates by survey round and urbanicity at the enumeration-area level. EA Count denotes the number of unique survey enumeration areas in each group. Reported values are percentages.}
\label{tab:urban_rural_summary}
\small
\setlength{\tabcolsep}{8pt}
\begin{tabular}{llccc}
\toprule
\textbf{Round} & \textbf{Urbanicity} & \textbf{EA Count} & \textbf{Piped water (\%)} & \textbf{Sewage system (\%)} \\
\midrule
R7 & Rural & 4683 & 34.3 & 12.6 \\
R7 & Urban & 4040 & 78.3 & 54.8 \\
R8 & Rural & 3326 & 33.9 & 10.9 \\
R8 & Urban & 2699 & 80.7 & 54.9 \\
R9 & Rural & 3504 & 31.0 & 9.9 \\
R9 & Urban & 3150 & 82.8 & 56.8 \\
\bottomrule
\end{tabular}
\end{table}

\FloatBarrier

\end{document}